\documentclass[sigconf,screen,nonacm]{acmart}
\usepackage{multirow}
\usepackage{booktabs}
\usepackage{graphicx}
\usepackage{enumitem}
\usepackage{natbib}

\usepackage{subcaption}
\usepackage{tcolorbox}
\usepackage{algorithm}
\usepackage{algorithmic}

\begin{document}

\title{CGC: Compositional Grounded Contrast for Fine-Grained Multi-Image Understanding}

\author{Lihao Zheng}
\affiliation{
  \institution{School of Computer Science and Technology, Hangzhou Dianzi University}
  \city{} \country{}
}

\author{Zhenwei Shao}
\affiliation{
  \institution{School of Computer Science and Technology, Hangzhou Dianzi University}
  \city{} \country{}
}

\author{Yu Zhou}
\affiliation{
  \institution{School of Computer Science and Technology, Hangzhou Dianzi University}
  \city{} \country{}
}

\author{Yan Yang}
\affiliation{
  \institution{School of Computer Science and Technology, Hangzhou Dianzi University}
  \city{} \country{}
}

\author{Xintian Shen}
\affiliation{
  \institution{Base Model, \\ Li Auto}
  \city{} \country{}
}

\author{Jiawei Chen}
\affiliation{
  \institution{Base Model, \\ Li Auto}
  \city{} \country{}
}

\author{Hao Ma}
\affiliation{
  \institution{Base Model, \\ Li Auto}
  \city{} \country{}
}

\author{Tao Wei}
\affiliation{
  \institution{Base Model, \\ Li Auto}
  \city{} \country{}
}

\begin{abstract}
Although Multimodal Large Language Models (MLLMs) have advanced rapidly, they still face notable challenges in fine-grained multi-image understanding, often exhibiting spatial hallucination, attention leakage, and failures in object constancy. In addition, existing approaches typically rely on expensive human annotations or large-scale chain-of-thought (CoT) data generation. We propose \textbf{Compositional Grounded Contrast} (abbr. \textbf{CGC}), a low-cost full framework for boosting fine-grained multi-image understanding of MLLMs. Built on existing single-image grounding annotations, CGC constructs compositional multi-image training instances through \textit{Inter-Image Contrast} and \textit{Intra-Image Contrast}, which introduce semantically decoupled distractor contexts for cross-image discrimination and correlated cross-view samples for object constancy, respectively. CGC further introduces a \textit{Rule-Based Spatial Reward} within the GRPO framework to improve source-image attribution, spatial alignment, and structured output validity under a \textit{Think-before-Grounding} paradigm. Experiments show that CGC achieves state-of-the-art results on fine-grained multi-image benchmarks, including MIG-Bench and VLM2-Bench. The learned multi-image understanding capability also transfers to broader multimodal understanding and reasoning tasks, yielding consistent gains over the Qwen3-VL-8B base model on MathVista (+2.90), MuirBench (+2.88), MMStar (+1.93), MMMU (+1.77), and BLINK (+1.69).
\end{abstract}



\begin{teaserfigure}
  \centering
    \includegraphics[trim=0mm 28mm 11mm 0mm, clip,width=\textwidth]{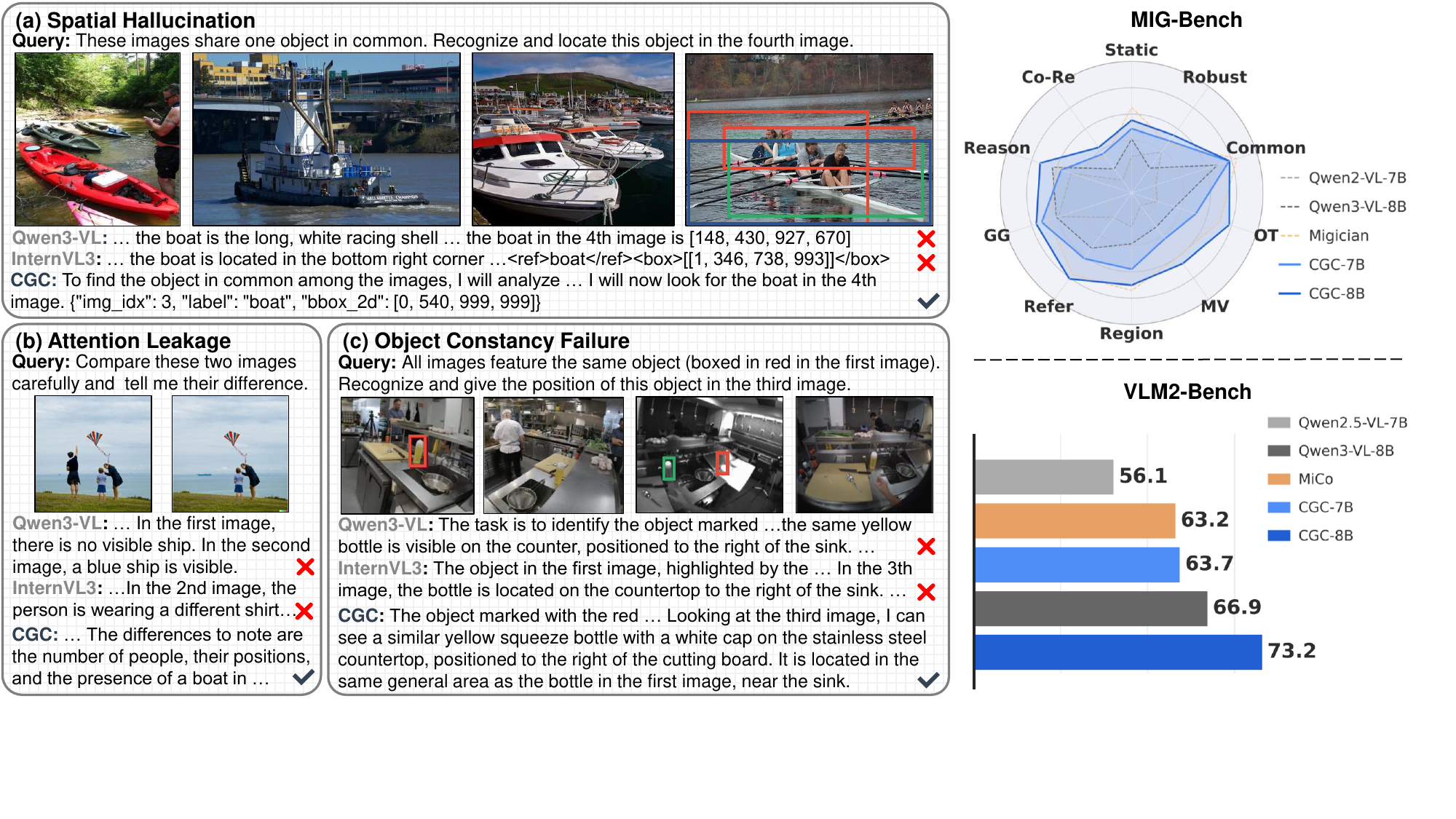}
    \caption{A qualitative and quantitative overview of the proposed CGC models. \textit{(Left)} Leading MLLMs often fail at fine-grained multi-image understanding primarily due to (a) spatial hallucination~\cite{hallusionbench, chen2024multi}, (b) attention leakage~\cite{park2025mitigating, qiaomultiple}, and (c) object constancy failure~\cite{CoLVA, assran2025v}, while CGC exhibits strong robustness in these cases. \textit{(Right)} CGC consistently outperforms strong baselines on diverse fine-grained multi-image understanding tasks on MIG-Bench~\cite{li2025migician} and VLM2-Bench~\cite{vlm2bench}.} 
    \Description{A qualitative and quantitative overview of the proposed CGC models. \textit{(Left)} Leading MLLMs often fail at fine-grained multi-image understanding primarily due to (a) spatial hallucination~\cite{hallusionbench, chen2024multi}, (b) attention leakage~\cite{park2025mitigating, qiaomultiple}, and (c) object constancy failure~\cite{CoLVA, assran2025v}, while CGC exhibits strong robustness in these cases. \textit{(Right)} CGC consistently outperforms strong baselines on diverse fine-grained multi-image understanding tasks on MIG-Bench~\cite{li2025migician} and VLM2-Bench~\cite{vlm2bench}.} 
  \label{fig:fig1}
\end{teaserfigure}

\maketitle

\section{Introduction}

Multimodal Large Language Models (MLLMs) have achieved strong performance on single-image perception and reasoning tasks~\cite{liu2023visual, achiam2023gpt}, yet they remain much weaker in multi-image tasks~\cite{wang2024muirbench, meng2024mmiu, liu2024mibench}. Fine-grained multi-image understanding requires the model not only to recognize and localize a target, but also to attribute that target to the correct source image under cross-image interference. As shown in Fig.~\ref{fig:fig1}, leading foundation models such as Qwen3-VL~\cite{yang2025qwen3} and InternVL3~\cite{zhu2025internvl3} often fail to maintain strict image-text attribution boundaries across heterogeneous image sequences.

This failure manifests in several ways. First, feature entanglement across images leads to \textit{cross-image attention leakage}~\cite{park2025mitigating, qiaomultiple}, where visual evidence from one image contaminates the interpretation of another (Fig.~\ref{fig:fig1}(b)). Second, without explicit coordinate-level grounding, models are prone to \textit{spatial hallucination}~\cite{hallusionbench, chen2024multi}, producing plausible but incorrect localizations (Fig.~\ref{fig:fig1}(a)). Third, under viewpoint or contextual variation, they often fail to preserve \textit{object constancy}~\cite{CoLVA, assran2025v}, losing track of the same entity across images or transformed views (Fig.~\ref{fig:fig1}(c)). Together, these issues reveal that weak low-level grounding remains a major bottleneck for fine-grained multi-image understanding.

To address these bottlenecks, recent methods have increasingly turned to Reinforcement Learning (RL) to enhance the visual capabilities of MLLMs. Specifically, Reinforcement Learning with Verifiable Rewards (RLVR) via algorithms like GRPO~\cite{shao2024deepseekmath} has shown great promise~\cite{liu2025visual, shen2025vlm}. For instance, grounding-oriented methods that optimize objective metrics such as bounding-box IoU have achieved notable gains in single-image scenarios~\cite{cao2025ground, gao2026diva}. However, extending these verifiable RL paradigms to fine-grained multi-image understanding remains largely unexplored. Existing approaches typically operate within single-image contexts and do not directly resolve cross-image referential ambiguity, where the model must first attribute the target to the correct source image before grounding it.\looseness=-1

In addition to methodological limitations, existing fine-grained multi-image methods are heavily constrained by data curation costs and weak supervision.
First, many approaches rely on costly cold-start supervised fine-tuning (SFT) with manual annotations (e.g., Migician~\cite{li2025migician}) or large-scale chain-of-thought (CoT) data generation. This not only increases curation overhead but may also introduce unstable or factually inconsistent training signals. Second, recent self-supervised or contrastive approaches (e.g., MiCo) mainly capture global image-level differences but lack explicit spatial supervision over source-image attribution and object location. As a result, they improve coarse cross-image discrimination without fully resolving referential ambiguity in fine-grained tasks.

To address these challenges, we propose \textbf{CGC} (\textbf{Compositional Grounded Contrast}), a low-cost full framework designed to improve fine-grained multi-image understanding by reinforcing a base MLLM to tackle compositional grounding tasks. The key idea of CGC is to use low-level spatial grounding as a verifiable stepping stone for high-level multi-image reasoning. 

By adopting a \textit{Think-before-Grounding} paradigm, CGC requires neither expensive multi-image human annotations nor LLM-based template design, data filtering, or large-scale CoT generation.
Specifically, our framework operates in two synergistic stages:

In the first stage, \textit{Automated Contrastive Data Synthesis} converts static single-image grounding annotations into challenging multi-image training datasets. This compositional process is implemented through two complementary mechanisms: \textit{Inter-Image Contrast}, which introduces semantically decoupled distractor contexts for cross-image discrimination, and \textit{Intra-Image Contrast}, which generates correlated cross-view samples to strengthen object constancy and spatial consistency.

In the second stage, CGC optimizes over the synthesized training instances with a \textit{Rule-Based Spatial Reward} under GRPO. Given a multi-image query, the model produces a structured response consisting of a reasoning trace followed by a JSON grounding output containing \texttt{img\_idx}, \texttt{label}, and \texttt{bbox\_2d}. The reward combines source-aware set-wise IoU with strict format validation, encouraging the model to attribute targets to the correct image while maintaining accurate localization and valid structured outputs.

Experiments show that CGC, built upon Qwen3-VL-8B and post-trained exclusively on synthesized data, achieves state-of-the-art performance on fine-grained multi-image benchmarks including MIG-Bench~\cite{li2025migician} and VLM2-Bench~\cite{vlm2bench}. Beyond these targeted gains, CGC also improves general multimodal reasoning and understanding on broader benchmarks, yielding consistent gains on MathVista (+2.90), MuirBench (+2.88), MMStar (+1.93), MMMU (+1.77), and BLINK (+1.69). These results suggest that stronger low-level grounding provides a highly transferable foundation for higher-level multimodal capabilities.

In summary, the main contributions of this paper are as follows:
\begin{itemize}[leftmargin=*, noitemsep, topsep=2pt]

    \item We propose an automated compositional synthesis pipeline that transforms existing single-image grounding annotations into challenging multi-image training instances without human annotation. By combining \textit{Inter-Image Contrast} to introduce semantically decoupled distractor contexts for cross-image discrimination and \textit{Intra-Image Contrast} to generate correlated cross-view samples for object constancy, this pipeline provides a scalable and cost-effective alternative to expensive human annotation and large-scale CoT data generation.

    \item We design a \textit{Rule-Based Spatial Reward} within the GRPO framework to improve the model's ability to perform source-image attribution, spatial alignment, and valid structured output generation.
    By combining source-aware set-wise IoU with strict format checks under a \textit{Think-before-Grounding} paradigm, the reward provides an effective optimization signal for fine-grained multi-image understanding.

    \item Built upon the Qwen3-VL-8B base model and post-trained exclusively on synthesized data, CGC achieves state-of-the-art results on fine-grained multi-image benchmarks including MIG-Bench and VLM2-Bench. Furthermore, the learned capabilities transfer effectively to broader multimodal reasoning tasks, yielding consistent gains on MathVista (+2.90), MuirBench (+2.88), MMStar (+1.93), MMMU (+1.77), and BLINK (+1.69).

\end{itemize}

\section{Related Work}
\subsection{Visual Grounding in MLLMs}
Recent Multimodal Large Language Models (MLLMs), such as Shikra~\cite{chen2023shikra}, Griffon~\cite{zhan2024griffon}, and Ferret~\cite{you2023ferret}, formulate visual grounding as a language generation problem by serializing spatial coordinates into text tokens. This line of work establishes grounding as a practical interface between language generation and object localization. To further improve grounding fidelity and reduce hallucination, subsequent approaches incorporate stronger supervision or optimization strategies, including preference learning~\cite{yu2024rlhf} and reinforcement learning for single-image perception~\cite{ma2025deepperception, liu2025visual}. More recently, several works have begun to extend grounding to multi-image tasks, including benchmark construction and post-training methods such as Migician~\cite{li2025migician}, MIRG-RL~\cite{zheng2025mirg}, and GeM-VG~\cite{zheng2026gem}. While these efforts demonstrate the importance of grounding in multi-image scenarios, they typically rely on manually curated multi-image annotations, cross-image question-answer pairs, or teacher-generated reasoning data. By contrast, our approach reuses existing single-image grounding annotations and recomposes them into verifiable multi-image training instances.

\subsection{Fine-Grained Multi-Image Understanding}
As MLLMs become increasingly capable on single-image tasks, recent research has shifted toward multi-image understanding, driven by benchmarks such as BLINK~\cite{fu2024blink}, MuirBench~\cite{wang2024muirbench}, and MMIU~\cite{meng2024mmiu}. These benchmarks highlight the challenge of aggregating evidence across multiple images, panels, or spatially separated visual regions. However, most existing approaches still focus on coarse-grained semantic comparison or high-level reasoning, without explicit mechanisms for source-image attribution and fine-grained spatial localization. SFT-based multi-image methods, including models such as Mantis~\cite{jiang2024mantis} and data resources such as OBELICS~\cite{laurenccon2023obelics}, improve multi-image adaptability but often rely on expensive manual curation and provide weak corrective signals under severe cross-image interference. In contrast, recent self-supervised or contrastive approaches such as MiCo~\cite{chen2025mico} improve cross-image discrimination by modeling global image-level similarities, but do not impose explicit spatial grounding constraints on source-image attribution and localization. More broadly, recent efforts on multimodal capability enhancement, including tool-integrated reasoning~\cite{chen2025mindwatcher}, training-free experience reuse~\cite{shen2026evolving}, search-agent evaluation~\cite{chen2026evaluating}, multi-stage post-training~\cite{mindgpt-4ov}, and large-scale multimodal system development~\cite{chen2026streamingclaw}, further demonstrate the growing interest in improving complex multimodal understanding. However, these approaches primarily target general reasoning, tool use, or overall model capability, rather than explicit cross-image grounding with verifiable source attribution. Motivated by this gap, we treat fine-grained multi-image understanding as a structured grounding problem over image index and spatial location.\looseness=-1

\subsection{Reinforcement Learning for MLLMs}
Rule-based reinforcement learning, including GRPO-style optimization~\cite{shao2024deepseekmath, guo2025deepseek}, has recently been adapted to a growing range of vision-language tasks. Existing visual RL methods broadly target either high-level multimodal reasoning~\cite{huang2025visionr1incentivizingreasoningcapability, deng2025openvlthinker, zhang2025r1} or fine-grained visual perception~\cite{shen2025vlm, yu2025perception}. Compared with free-form reasoning tasks, grounding-oriented tasks are particularly suitable for rule-based RL because they provide objectively verifiable reward signals. Recent grounding-oriented RL works exploit this property using training-free or rule-based rewards, including difficulty-adaptive reward design in DIVA-GRPO~\cite{gao2026diva}, scale-bias mitigation in Ground-R1~\cite{cao2025ground}, and explicit visual reasoning traces in DIP-R1~\cite{park2025dip}. 
However, these methods remain largely confined to single-image tasks, where the optimization target is not source-image attribution under cross-image interference. In contrast, our framework extends rule-based visual RL to fine-grained multi-image understanding through source-aware spatial rewards and training signals derived from recomposed single-image grounding annotations.

\begin{figure*}[htbp]
    \centering
    \includegraphics[trim=4mm 1mm 5mm 5mm, clip, width=\textwidth]{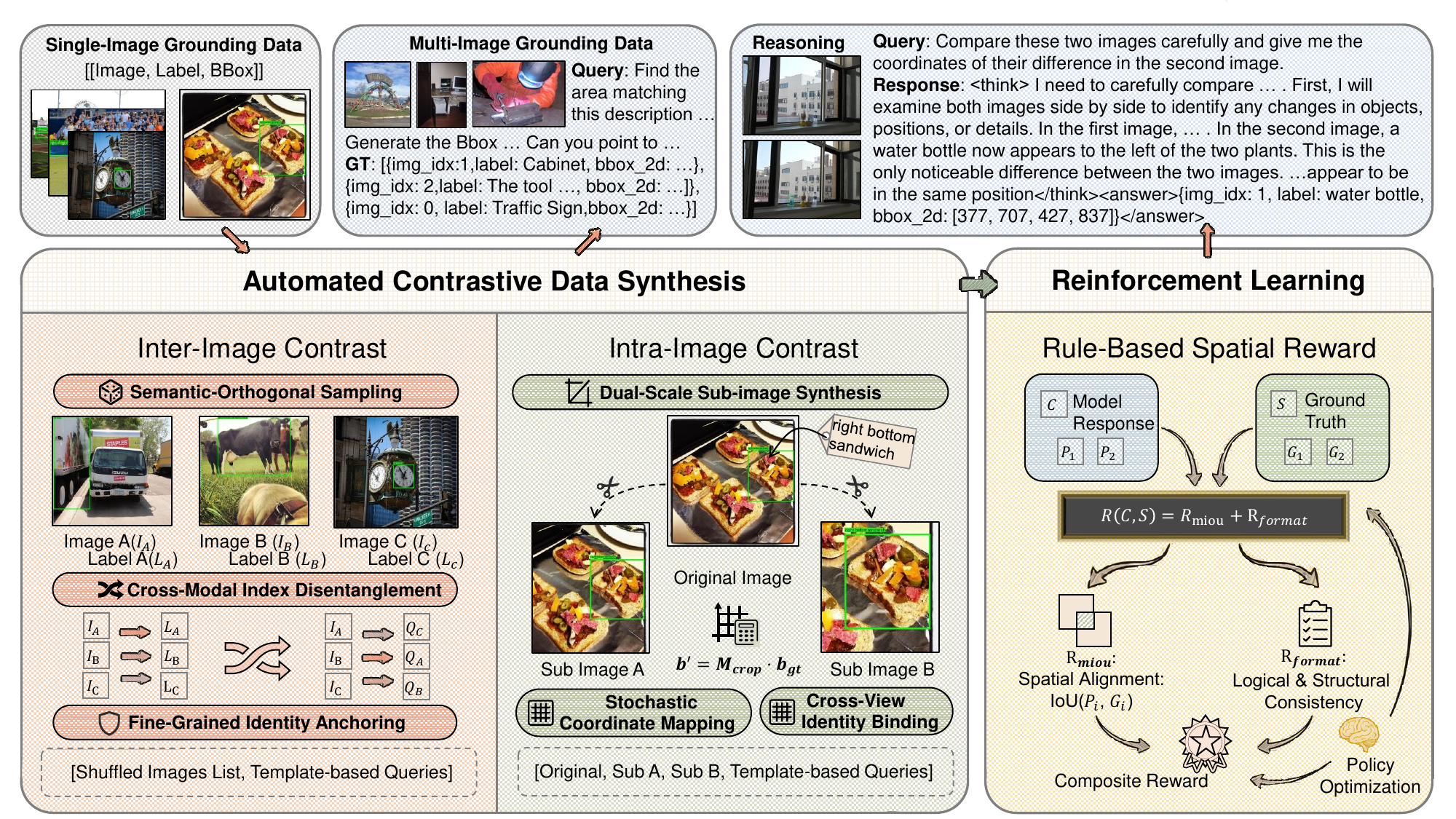}
    \caption{Overview of the proposed \textbf{CGC} framework. Starting from single-image grounding annotations, CGC automatically constructs multi-image training instances through \textit{Automated Contrastive Data Synthesis}, which includes \textit{Inter-Image Contrast} for decoupled distractor contexts and \textit{Intra-Image Contrast} for correlated cross-view grounding. Under a \textit{Think-before-Grounding} paradigm, the model first produces a reasoning trace and then outputs a structured grounding prediction. Finally, \textit{Rule-Based Spatial Reward} optimization under GRPO jointly improves source-image attribution, spatial alignment, and output validity.}    \Description{A diagram showing the CGC framework.}
    \label{fig:0227}
\end{figure*}

\section{Method}
\label{sec:method}

We propose \textbf{CGC} (\textbf{Compositional Grounded Contrast}), a low-cost full framework for fine-grained multi-image understanding in MLLMs. Starting from existing single-image grounding annotations, CGC automatically constructs compositional multi-image data and optimizes the model with rule-based reinforcement learning.

\subsection{Task Formulation}
Let $\pi_\theta$ denote an MLLM parameterized by $\theta$. Given a multi-image input sequence $\mathbf{I} = [I_1, I_2, \dots, I_K]$ and a textual query $q$, the model generates a response $o \sim \pi_\theta(o \mid \mathbf{I}, q)$. The query $q$ is instantiated from a template pool $\mathcal{T}$, where each template $\tau \in \mathcal{T}$ defines a textual pattern for referring to targets across images. Under the \textit{Think-before-Grounding} paradigm, the response is decomposed as $o = (t, C)$, where $t$ is a reasoning trace and $C$ is the final structured grounding prediction. We represent $C$ as a set of target tuples $C = \{p_1, \dots, p_{\hat{M}}\}$, where each prediction $p_i = (\hat{k}_i, \hat{y}_i, \hat{b}_i)$ consists of an image index $\hat{k}_i \in \{0, \dots, K-1\}$, a target label $\hat{y}_i$, and a bounding box $\hat{b}_i = [\hat{x}_{1i}, \hat{y}_{1i}, \hat{x}_{2i}, \hat{y}_{2i}]$. In practice, $C$ is serialized as a JSON-style output with the fields \texttt{img\_idx}, \texttt{label}, and \texttt{bbox\_2d}. The corresponding ground-truth target set is denoted by $S = \{g_1, \dots, g_M\}$, where each $g_j = (k_j, y_j, b_j)$ specifies the source image, target label, and bounding box of a target.

\subsection{Inter-Image Contrast}
\label{sec:inter_image}

Inter-Image Contrast constructs distractor-rich multi-image training samples by combining semantically decoupled images within the same input. Because these images typically depict unrelated scenes and objects, they introduce strong cross-image interference and encourage explicit source-image attribution and target disambiguation, rather than allowing the model to rely on superficial similarity patterns or default positional correspondences.

\noindent\textbf{Semantic-Orthogonal Sampling.}
We sample $K$ annotated images without replacement from the grounding dataset $\mathcal{D}$, where each image is associated with target labels and bounding-box annotations, to form a multi-image input sequence $\mathbf{I} = [I_1, I_2, \dots, I_K]$. Because these images are independently drawn, they usually contain unrelated visual content, yielding semantically decoupled contexts. This reduces shortcut learning based on superficial cross-image similarity and encourages explicit target disambiguation.

\noindent\textbf{Cross-Modal Index Disentanglement.}
We further shuffle the image order, together with their associated references, as part of the construction process for each training sample. 
By making use of a template pool $\mathcal{T}$, we instantiate diverse textual queries $q$ from the source annotations, where target labels provide the semantic referents and bounding boxes define the grounding targets.
This procedure breaks the default positional correspondence between textual references and visual inputs, thereby reducing the model's tendency to rely on sequence bias and consequently promoting explicit mapping from the query to image indices.

\noindent\textbf{Fine-Grained Identity Anchoring.}
Each target in the synthesized supervision is explicitly anchored by its source image index, semantic label, and spatial location. As a result, the model must identify and localize the correct object in the designated image $I_k$ while treating all remaining images $I_{j \ne k}$ as distractors, thereby strengthening cross-image discrimination.

\subsection{Intra-Image Contrast}
\label{sec:intra_image}
Intra-Image Contrast constructs correlated multi-image data from a single source image by synthesizing multiple views of the same target. This branch provides identity-consistent supervision across variations in crop scale, surrounding context, and spatial layout.

\noindent\textbf{Dual-Scale Sub-image Synthesis.}
Given a source image $I_{\text{src}}$ and a target annotation $g = (y, b)$, where $y$ denotes the semantic target label and $b$ denotes the target bounding box, we generate two object-centered sub-images with different crop scales around the same target. A smaller crop, denoted as the \emph{Focus View}, preserves fine-grained target appearance and local visual details, while a larger crop, denoted as the \emph{Context View}, retains richer surrounding context and broader spatial cues. In practice, the crop ratio is sampled from $1.2\times$--$1.5\times$ object scale for the Focus View and from $1.8\times$--$2.5\times$ object scale for the Context View. These two views provide complementary evidence for grounding the same target.

\noindent\textbf{Stochastic Coordinate Mapping.}
We apply randomized object-centered cropping to diversify the synthesized views while ensuring full target visibility. This process creates varied target layouts and reduces shortcut learning from fixed spatial patterns. For each crop, the original box $b$ is deterministically mapped into the local coordinate system to obtain a transformed box $b'$. The resulting labels remain geometrically consistent across views and provide reliable supervision for cross-view localization.

\noindent\textbf{Cross-View Identity Binding.}
We combine the synthesized views, together with the original image, into a correlated multi-image input and shuffle their order before query construction. Using a template pool $\mathcal{T}$, we construct co-referential queries $q$ that refer to the same target identity across views. The model must then identify the same object under different crops and scales, and predict its corresponding image index and bounding box in each view.  

\subsection{Data Mixing}
CGC mixes Inter-Image Contrast and Intra-Image Contrast samples with a fixed 1:1 ratio. This design combines two complementary mechanisms for synthesized multi-image construction. Inter-Image Contrast strengthens source-image attribution and cross-image discrimination by introducing semantically decoupled distractor contexts, whereas Intra-Image Contrast improves object constancy by exposing the model to correlated views of the same target. Together, the two mechanisms cover complementary aspects of fine-grained multi-image grounding, including source-image attribution, cross-view identity consistency, and precise localization.

\subsection{GRPO via Rule-Based Spatial Reward}
We optimize CGC with Group Relative Policy Optimization (GRPO) to improve structured spatial grounding in multi-image tasks. Given an input $(\mathbf{I}, q)$, the model generates a response $o = (t, C)$ under the \textit{Think-before-Grounding} paradigm. We parse the final structured grounding prediction $C$ and compute a verifiable reward based on spatial correctness and output validity. For a parsed prediction $C$ and ground truth $S$, the total reward is defined as:
\[
R(C, S) = R_{\text{miou}}(C, S) + R_{\text{format}}(C),
\]
where $R_{\text{miou}}$ measures source-aware spatial alignment and $R_{\text{format}}$ enforces valid structured output.

\noindent\textbf{Source-Aware Set-Wise IoU Reward ($R_{\text{miou}}$).}
To evaluate unordered multi-object predictions, we formulate coordinate evaluation as a source-aware set matching problem over image-indexed target groups.
Specifically, we parse the structured prediction $C$ into a prediction set $\mathcal{P}$ and represent the ground truth $S$ as a target set $\mathcal{G}$. We then partition $\mathcal{P}$ and $\mathcal{G}$ into disjoint subsets $\mathcal{P}_k$ and $\mathcal{G}_k$ according to image index $k$. For each image-index subset, we perform optimal bipartite matching to maximize the total Intersection over Union (IoU) between predicted and ground-truth boxes. Because matching is restricted within the same image index, predictions assigned to the wrong image are naturally penalized, which directly suppresses cross-image attribution errors.

To penalize cardinality mismatches such as redundant predictions, we normalize the matched IoU sum by the larger set size:
\[
R_{\text{miou}} =
\frac{
\sum_{(b_p, b_g)\in \mathrm{Matches}} \operatorname{IoU}(b_p, b_g)
}{
\max(|\mathcal{P}|, |\mathcal{G}|)
}.
\]
This reward is maximized only when the model predicts the correct image index, number of targets, and box locations.

\noindent\textbf{Format and Validity Reward ($R_{\text{format}}$).}
We use a binary reward:
\[
R_{\text{format}}(C) \in \{0, 1\}.
\]
A positive reward is assigned only when the output satisfies both \emph{response-format} and \emph{structural validity} constraints. 

For response-format, the full response must follow the \textit{Think-before-Grounding} paradigm, consisting of a tagged reasoning segment followed by a tagged answer segment, and the final answer must be parseable as a structured grounding output.
In addition, each predicted target must contain exactly the required fields for image index, semantic label, and bounding box, with no missing or malformed fields. For structural validity, each predicted box $[x_1, y_1, x_2, y_2]$ must form a non-degenerate rectangle satisfying $x_2 > x_1$ and $y_2 > y_1$, and all coordinates must lie within $[0, 1000)$.

\noindent\textbf{Training Objective.}
Given an input $(\mathbf{I}, q)$, the policy $\pi_{\theta}$ samples a group of $G$ responses $\{o_1, \dots, o_G\}$. Each response $o_i$ is parsed into a structured prediction $C_i$, which receives a scalar reward $R_i = R(C_i, S)$. GRPO computes the normalized group advantage as:
\[
A_i =
\frac{
R_i - \operatorname{mean}(\{R_1, \dots, R_G\})
}{
\operatorname{std}(\{R_1, \dots, R_G\})
}.
\]

We define the probability ratio as:
\[
r_i(\theta)=\frac{\pi_{\theta}(o_i \mid \mathbf{I}, q)}{\pi_{\text{old}}(o_i \mid \mathbf{I}, q)}.
\]

The optimization objective is:
\[
\begin{aligned}
\mathcal{J}(\theta)
= \mathbb{E}_{(\mathbf{I}, q),\{o_i\}}
\Biggl[
\frac{1}{G}\sum_{i=1}^{G}
\Bigl(
&\min \bigl(
r_i(\theta)A_i,\,
\operatorname{clip}(r_i(\theta), 1-\epsilon, 1+\epsilon)A_i
\bigr)
\\
&- \beta \,\mathbb{D}_{KL}(\pi_{\theta}\,\|\,\pi_{\text{ref}})
\Bigr)
\Biggr].
\end{aligned}
\]

Here, $\epsilon$ is the clipping hyperparameter and $\mathbb{D}_{KL}$ is the KL regularization term that constrains the updated policy to remain close to the reference model $\pi_{\text{ref}}$, thereby reducing catastrophic forgetting. We compute the KL penalty as:
\[
\mathbb{D}_{KL}(\pi_{\theta}\,\|\,\pi_{\text{ref}})
=
\frac{\pi_{\text{ref}}(o_i \mid \mathbf{I}, q)}{\pi_{\theta}(o_i \mid \mathbf{I}, q)}
-
\log
\frac{\pi_{\text{ref}}(o_i \mid \mathbf{I}, q)}{\pi_{\theta}(o_i \mid \mathbf{I}, q)}
-
1.
\]

By maximizing $\mathcal{J}(\theta)$, the model improves the accuracy, consistency, and structural reliability of fine-grained grounding and understanding over multiple images.

\begin{table*}[tb]
\centering
\caption{Performance comparison on MIG-Bench. Best non-human results are shown in \textbf{bold}.}
\label{tab:mig_bench}
\resizebox{\textwidth}{!}{
\begin{tabular}{lcccccccccccc}
\toprule

\multirow{3}{*}{\textbf{Models}} & \multirow{3}{*}{\textbf{Base MLLM}} & \multicolumn{3}{c}{\textbf{Spontaneous Grounding}} & \multicolumn{7}{c}{\textbf{Referential Grounding}} & \multirow{3}{*}{\textbf{AVE}} \\
\cmidrule(lr){3-5} \cmidrule(lr){6-12}
& & \multicolumn{2}{c}{\textbf{Difference}} & \textbf{Similarity} & \multicolumn{4}{c}{\textbf{Visual Reference}} & \textbf{Textual} & \multicolumn{2}{c}{\textbf{Visual+Textual}} & \\
\cmidrule(lr){3-4} \cmidrule(lr){5-5} \cmidrule(lr){6-9} \cmidrule(lr){10-10} \cmidrule(lr){11-12}
& & \textbf{Static} & \textbf{Robust} & \textbf{Common} & \textbf{OT} & \textbf{MV} & \textbf{Region} & \textbf{Refer} & \textbf{GG} & \textbf{Reason} & \textbf{Co-Re} & \\
\midrule
Human & - & 99.50 & 97.87 & 98.00 & 100.00 & 96.88 & 100.00 & 98.99 & 91.06 & 92.08 & 97.44 & 97.18 \\
\midrule
\multicolumn{13}{l}{\small\textit{Large-Scale Models ($\geq$70B)}} \\
LLaVA-OV-72B~\cite{li2024llava} & - & 13.26 & 5.34 & 26.84 & 12.91 & 7.64 & 2.14 & 17.83 & 21.60 & 11.88 & 8.55 & 13.65 \\
InternVL2-76B~\cite{team2024internvl2} & - & 15.91 & 10.64 & 36.40 & 30.73 & 20.83 & 5.74 & 46.46 & 41.28 & 32.67 & 26.50 & 26.72 \\
InternVL3-78B~\cite{zhu2025internvl3} & - & 10.04 & 9.57 & 24.12 & 27.08 & 14.58 & 10.44 & 50.51 & 38.08 & 45.54 & 17.09 & 24.71 \\
Qwen2-VL-72B~\cite{wang2024qwen2} & - & 46.12 & 46.81 & 64.46 & 26.73 & 22.57 & 18.62 & 33.33 & 62.53 & 50.50 & 17.09 & 38.88 \\
Qwen2.5-VL-72B~\cite{bai2025qwen2} & - & 43.75 & 46.81 & 69.98 & 34.32 & 29.17 & 8.31 & 62.63 & 59.92 & 66.34 & 41.03 & 46.23 \\
\midrule
\multicolumn{13}{l}{\small\textit{7 $\sim$8B\,Models}} \\
Mantis-8B~\cite{jiang2024mantis} & - & 1.52 & 0.00 & 3.31 & 12.18 & 2.08 & 1.00 & 1.01 & 10.02 & 0.00 & 0.85 & 3.20 \\
LLaVA-OV-7B~\cite{li2024llava} & - & 6.06 & 3.19 & 3.43 & 0.18 & 1.04 & 1.08 & 9.09 & 15.43 & 6.93 & 0.85 & 4.73 \\
MiniCPM2.6-8B~\cite{yao2024minicpm} & - & 14.58 & 2.13 & 14.34 & 9.82 & 6.25 & 1.75 & 11.11 & 10.02 & 2.97 & 2.56 & 7.55 \\
mPLUG-Owl3-8B~\cite{ye2024mplug} & - & 18.56 & 6.38 & 34.93 & 8.55 & 7.64 & 2.41 & 7.07 & 22.85 & 9.09 & 5.98 & 12.35 \\
InternVL2-8B~\cite{team2024internvl2} & - & 6.92 & 7.45 & 25.49 & 20.73 & 9.72 & 3.49 & 28.28 & 30.26 & 17.82 & 9.40 & 15.96 \\
InternVL3-8B~\cite{zhu2025internvl3} & - & 23.67 & 14.89 & 47.99 & 14.84 & 6.94 & 12.13 & 7.07 & 34.87 & 16.83 & 2.56 & 18.18 \\
Qwen2-VL-7B~\cite{wang2024qwen2} & - & 27.84 & 38.30 & 19.36 & 20.73 & 11.81 & 25.95 & 23.23 & 58.52 & 48.51 & 11.97 & 28.62 \\
Qwen2.5-VL-7B~\cite{bai2025qwen2} & - & 9.57 & 21.59 & 22.42 & 10.58 & 5.88 & 1.25 & 10.00 & 33.47 & 2.97 & 2.60 & 12.03 \\
Qwen3-VL-8B~\cite{yang2025qwen3} & - & 41.10 & 23.40 & 68.63 & 34.59 & 31.94 & 38.73 & 52.00 & 60.16 & 63.37 & 22.65 & 43.66 \\
Visual-RFT-7B~\cite{liu2025visual} & Qwen2-VL & 33.95 & 31.29 & 50.00 & 15.36 & 12.21 & 14.98 & 24.25 & 62.72 & 41.89 & 10.40 & 29.71 \\
Migician-7B~\cite{li2025migician} & Qwen2-VL & \textbf{65.15} & 46.81 & \textbf{84.19} & 70.73 & 60.07 & \textbf{74.31} & 76.77 & 66.53 & 59.41 & 34.19 & 63.82 \\
\midrule
\textbf{CGC-7B (ours)} & Qwen2-VL & 48.86 & 48.94 & 77.21 & 51.45 & 59.72 & 58.02 & 61.62 & 71.94 & 56.44 & 36.75 & 55.50 \\
\textbf{CGC-8B (ours)} & Qwen3-VL & 55.30 & \textbf{54.26} & 78.31 & \textbf{78.18} & \textbf{66.32} & 70.16 & \textbf{80.81} & \textbf{76.35} & \textbf{73.27} & \textbf{42.74} & \textbf{67.57} \\
\bottomrule
\end{tabular}
}
\end{table*}

\section{Experiments}

\subsection{Experimental Setup}

\textbf{Training Data Construction.}
To ensure reliable visual feature extraction, we retain only images with a minimum edge length of 640 pixels. We construct the base single-image grounding set from the training and validation splits of five open-source datasets, including RefCOCO~\cite{yu2016modeling}, SODA~\cite{yuan2023small}, LISA~\cite{lai2024lisa}, OmniLabel~\cite{schulter2023omnilabel}, and VAW~\cite{pham2021learning}, while strictly excluding all test sets to prevent data leakage. This results in approximately 72,000 single-image grounding instances. Using the automated contrastive data synthesis procedures introduced in Sec.~\ref{sec:inter_image} and Sec.~\ref{sec:intra_image}, we convert these single-image samples into approximately 36,000 multi-image training instances. The final reinforcement learning dataset uses a 1:1 mixture of inter-image contrast and intra-image contrast samples.

\noindent\textbf{Implementation Details.}
We train CGC with GRPO, assigning equal weights to the source-aware set-wise IoU reward ($R_{\text{miou}}$) and the format compliance reward ($R_{\text{format}}$).
The KL penalty coefficient $\beta$ is set to 0.01. Following standard practice, we skip optimization steps when all rollout candidates within a group receive identical rewards. We use a cosine learning rate scheduler with a warmup ratio of 0.05 and a peak learning rate of $1\times10^{-6}$. The global batch size is 128, and the policy samples $G=8$ candidate responses for each training query. We do not use any SFT stage, multi-image SFT data, or teacher-generated reasoning traces during training.

\noindent\textbf{Models and Benchmarks.}
We instantiate CGC on Qwen2-VL-7B-Instruct, Qwen2.5-VL-7B-Instruct, and Qwen3-VL-8B-Instruct as the base MLLMs. For fair comparison with benchmark-specific strong baselines, we use the Qwen2-VL-based variant on MIG-Bench to match Migician, the Qwen2.5-VL-based variant on VLM2-Bench and general vision benchmarks to match MiCo, and the Qwen3-VL-based variant as our 8B model. We evaluate fine-grained multi-image understanding on MIG-Bench and VLM2-Bench, and further assess broader multimodal understanding and reasoning on a diverse set of benchmarks, including BLINK, MMMU, MathVista, MMStar, HallusionBench, and MuirBench.

\subsection{Main Results}

Table~\ref{tab:mig_bench} and Table~\ref{tab:vlm2_bench} report the main results on MIG-Bench and VLM2-Bench. Overall, CGC achieves state-of-the-art performance across both benchmarks, suggesting that explicit spatial grounding provides an effective inductive bias for cross-image discrimination, referential grounding, and correspondence reasoning in complex multi-image tasks.

\noindent\textbf{Results on MIG-Bench.}
MIG-Bench evaluates fine-grained multi-image grounding from two complementary perspectives: \textit{Spontaneous Grounding}, which requires discovering implicit cross-image relations without explicit references, and \textit{Referential Grounding}, which tests grounding under multimodal references.

For a controlled comparison with Migician, we use the same backbone, Qwen2-VL-7B. As shown in Table~\ref{tab:mig_bench}, CGC-7B, obtained purely through GRPO-based post-training, achieves an average score of 55.50, substantially improving over the Qwen2-VL-7B base model (28.62) by 26.88 points. By contrast, Migician reaches 63.82 but relies on 630K multi-image SFT samples. This matched-backbone comparison suggests that strong multi-image grounding capability can be substantially improved without relying on large-scale, carefully curated multi-image SFT data.

CGC-8B further achieves the best overall average score of 67.57. It improves over its base model Qwen3-VL-8B (43.66) by 23.91 points, surpasses Migician (63.82), and also outperforms substantially larger general-purpose MLLMs, including Qwen2.5-VL-72B (46.23) and InternVL2-76B (26.72). The gains are especially pronounced on challenging cross-image alignment tasks. In the \texttt{Robust} task, CGC-8B improves from 23.40 to 54.26 over its base model. In \texttt{MV} and \texttt{Co-Re}, which emphasize cross-view consistency and region-level correspondence, CGC-8B reaches 66.32 and 42.74, compared with 31.94 and 22.65 for Qwen3-VL-8B. These improvements are consistent with the roles of \textit{Inter-Image Contrast} and \textit{Intra-Image Contrast}.

\begin{table*}[tb]
\centering
\caption{Performance comparison on VLM2-Bench. Best non-human results are shown in \textbf{bold}.}
\label{tab:vlm2_bench}
\resizebox{\textwidth}{!}{
\begin{tabular}{l c ccccccccc}
\toprule
\multirow{2}{*}{\textbf{Models}} & \multirow{2}{*}{\textbf{Base MLLM}} & \multicolumn{2}{c}{\textbf{General}} & \multicolumn{3}{c}{\textbf{Object}} & \multicolumn{3}{c}{\textbf{Person}} & \multirow{2}{*}{\textbf{AVE}} \\
\cmidrule(lr){3-4} \cmidrule(lr){5-7} \cmidrule(lr){8-10}
& & \textbf{Mat} & \textbf{Trk} & \textbf{Cnt (Obj)} & \textbf{Cpr (Obj)} & \textbf{Grp (Obj)} & \textbf{Cnt (Per)} & \textbf{Cpr (Per)} & \textbf{Grp (Per)} & \\
\midrule
Human-Level & - & 95.06 & 98.11 & 94.23 & 96.02 & 91.29 & 92.87 & 97.08 & 91.17 & 94.48 \\
\midrule
LLaVA-OV-7B~\cite{li2024llava} & - & 16.60 & 13.70 & 56.17 & 47.22 & 27.50 & 46.67 & 62.00 & 37.00 & 38.36 \\
LLaVA-Video-7B~\cite{llavavideo} & - & 18.53 & 12.79 & 62.47 & 54.72 & 28.50 & 66.91 & 62.00 & 25.00 & 41.37 \\
LongVQA-7B~\cite{longva} & - & 14.29 & 12.98 & 49.47 & 46.53 & 29.00 & 41.56 & 58.00 & 25.00 & 34.60 \\
mPLUG-Owl2-7B~\cite{mplugowl2} & - & 17.37 & 18.26 & 62.97 & 49.17 & 31.00 & 58.06 & 63.00 & 29.00 & 41.10 \\
InternVL2.5-8B~\cite{chen2024expanding} & - & 21.24 & 26.53 & 55.23 & 53.33 & 46.50 & 60.00 & 51.50 & 52.00 & 45.79 \\
InternVL2.5-26B~\cite{chen2024expanding} & - & 30.50 & 30.59 & 51.48 & 43.33 & 52.50 & 59.67 & 59.50 & 61.25 & 48.60 \\
InternVL3-8B~\cite{zhu2025internvl3} & - & 23.94 & 29.68 & 64.17 & 61.67 & 39.50 & 55.00 & 63.00 & 55.00 & 49.00 \\
Qwen2-VL-7B~\cite{wang2024qwen2} & - & 18.07 & 19.18 & 61.84 & 68.08 & 37.50 & 67.92 & 72.00 & 47.00 & 48.95 \\
Qwen2.5-VL-7B~\cite{bai2025qwen2} & - & 35.91 & 43.38 & 41.72 & 71.39 & 47.50 & 59.76 & 80.00 & 69.00 & 56.08 \\
Qwen3-VL-8B~\cite{yang2025qwen3} & - & 27.03 & 48.86 & 66.39 & 80.00 & 70.50 & 73.33 & 80.00 & 88.00 & 66.76 \\
\midrule
MM-Eureka-7B~\cite{meng2025mm} & Qwen2.5-VL & 55.60 & 47.03 & 52.50 & 74.10 & 54.00 & 60.00 & 77.50 & 51.00 & 58.97 \\
NoisyRollout-7B~\cite{liu2025noisyrollout} & Qwen2.5-VL & 40.93 & 43.83 & 50.83 & 63.33 & 34.50 & 63.33 & 70.50 & 47.00 & 51.78 \\
ThinkLite-VL-7B~\cite{ThinkLite} & Qwen2.5-VL & 40.45 & 46.58 & 62.50 & 75.56 & 49.50 & 62.50 & 77.50 & 51.00 & 58.20 \\
VLAA-Thinker-7B~\cite{VLAAThinker} & Qwen2.5-VL & 47.49 & 63.03 & 61.40 & 72.20 & 55.00 & 57.50 & 71.00 & 51.00 & 59.83 \\
MiCo-7B~\cite{chen2025mico} & Qwen2.5-VL & \textbf{57.14} & \textbf{67.12} & 56.67 & 81.94 & 58.00 & 57.50 & 65.00 & 62.00 & 63.17 \\
\midrule
\textbf{CGC-7B (ours)} & Qwen2.5-VL & 42.11 & 50.22 & 61.39 & 72.36 & 58.50 & 64.17 & 82.00 & 79.00 & 63.72 \\
\textbf{CGC-8B (ours)} & Qwen3-VL & 48.85 & 56.14 & \textbf{70.28} & \textbf{84.72} & \textbf{76.00} & \textbf{77.50} & \textbf{83.00} & \textbf{94.00} & \textbf{73.81} \\
\bottomrule
\end{tabular}
}
\end{table*}

\noindent\textbf{Results on VLM2-Bench.}
VLM2-Bench is a benchmark designed to assess whether MLLMs can visually link matching cues, focusing on cross-image matching and structural reasoning in the \texttt{General}, \texttt{Object}, and \texttt{Person} tracks, covering representative sub-tasks such as matching, tracking, counting, comparison, and grouping.

For a controlled comparison with MiCo, we instantiate CGC on the same Qwen2.5-VL-7B backbone. As shown in Table~\ref{tab:vlm2_bench}, CGC-7B achieves a slightly higher overall score than MiCo (63.72 vs.\ 63.17). The advantage is more evident on structurally demanding person-centric sub-tasks, including person counting (64.17 vs.\ 57.50), person comparison (82.00 vs.\ 65.00), and person grouping (79.00 vs.\ 62.00). These results suggest that our grounded data construction and spatially structured RL formulation are particularly beneficial for fine-grained multi-image understanding.

CGC-8B further establishes the state of the art on VLM2-Bench with an average score of 73.81, outperforming its base model Qwen3-VL-8B (66.76) by 6.90 points. It achieves the best non-human results on six out of eight sub-tasks, including object counting, object comparison, object grouping, person counting, person comparison, and person grouping. The gains are especially pronounced on person-centric tasks, where CGC-8B reaches 77.50 on person counting and 94.00 on person grouping, substantially surpassing MiCo.

Taken together, the results on both benchmarks show that CGC improves source-aware grounding, cross-image correspondence, and fine-grained reasoning in complex multi-image tasks.

\subsection{Generalization to Multimodal Reasoning}

We further evaluate whether CGC transfers beyond dedicated multi-image grounding benchmarks to broader multimodal understanding and reasoning tasks. Table~\ref{tab:general_vision} reports results on BLINK, HallusionBench, MathVista, MMMU, MMStar, and MuirBench. Although CGC is post-trained only on synthesized grounding data without task-specific QA supervision, it consistently generalizes to these broader benchmarks.

On the 7B scale, CGC achieves an average score of 63.41, outperforming several models trained with RL on the same backbone, including MiCo (62.61), MM-Eureka (62.54), NoisyRollout (62.36), and VLAA-Thinker (62.29). This result is noteworthy because CGC was not directly optimized for these downstream tasks 

and does not use any SFT stage or CoT supervision; instead, the gains emerge from GRPO post-training on synthesized grounding data alone.

At the 8B scale, CGC achieves the best overall average score of 71.65, improving over its base model Qwen3-VL-8B (69.74) by 1.91 points. CGC-8B obtains the best performance on BLINK, MathVista, MMMU, MMStar, and MuirBench, and remains competitive on HallusionBench. The strongest gains appear on MathVista (+2.90), MMMU (+1.77), and MuirBench (+2.88), indicating that grounded multi-image post-training is particularly helpful for tasks requiring structured evidence composition and spatially sensitive reasoning. It also improves HallusionBench from 71.40 to 72.56 and BLINK from 68.33 to 70.02, further suggesting that explicit low-level anchoring helps suppress unsupported visual claims and promotes more reliable evidence aggregation in visually complex tasks.

Overall, these results show that CGC improves not only dedicated multi-image grounding benchmarks but also a broader range of multimodal reasoning tasks. They further suggest that grounding-oriented post-training on low-cost synthesized data can induce general-purpose gains in visual evidence composition, structured comparison, and spatially sensitive reasoning.

\begin{table*}[tb]
\centering
\caption{Performance on general vision benchmarks. Best non-human results are shown in \textbf{bold}.}
\label{tab:general_vision}
\resizebox{0.9\linewidth}{!}{
\begin{tabular}{l c ccccccc}
\toprule
\textbf{Models} & \textbf{Base MLLM} & \textbf{BLINK} & \textbf{Hallusion} & \textbf{MathVista} & \textbf{MMMU} & \textbf{MMStar} & \textbf{MuirBench} & \textbf{AVE} \\
\midrule
InternVL3-8B~\cite{zhu2025internvl3} & - & 55.81 & 66.98 & 67.30 & 56.86 & 68.00 & 55.04 & 61.67 \\
Qwen2-VL-7B~\cite{wang2024qwen2} & - & 52.40 & 68.98 & 61.20 & 51.33 & 60.93 & 39.88 & 55.79 \\
Qwen2.5-VL-7B~\cite{bai2025qwen2} & - & 55.54 & 69.50 & 67.10 & 54.11 & 64.06 & 58.43 & 61.46 \\
Qwen3-VL-8B~\cite{yang2025qwen3} & - & 68.33 & 71.40 & 75.30 & 69.18 & 70.67 & 63.54 & 69.74 \\
\midrule
Migician-7B~\cite{li2025migician} & Qwen2-VL & 51.53 & 66.56 & 60.20 & 49.24 & 59.40 & 53.69 & 56.77 \\
MM-Eureka-7B~\cite{meng2025mm} & Qwen2.5-VL & 54.39 & 68.45 & 72.00 & 54.11 & 65.73 & 60.57 & 62.54 \\
NoisyRollout-7B~\cite{liu2025noisyrollout} & Qwen2.5-VL & 56.07 & 66.66 & 71.60 & 54.55 & 65.66 & 59.61 & 62.36 \\
VLAA-Thinker-7B~\cite{VLAAThinker} & Qwen2.5-VL & 54.81 & 69.08 & 70.80 & 54.44 & 63.60 & 61.00 & 62.29 \\
ThinkLite-VL-7B~\cite{ThinkLite} & Qwen2.5-VL & 55.81 & \textbf{72.97} & 71.89 & 53.55 & 66.80 & 57.62 & 63.11 \\
MiCo-7B~\cite{chen2025mico} & Qwen2.5-VL & 57.23 & 69.61 & 67.90 & 54.77 & 65.60 & 60.53 & 62.61 \\
\midrule
\textbf{CGC-7B (ours)} & Qwen2.5-VL & 56.23 & 72.45 & 70.90 & 55.71 & 64.20 & 60.88 & 63.41 \\
\textbf{CGC-8B (ours)} & Qwen3-VL & \textbf{70.02} & 72.56 & \textbf{78.20} & \textbf{70.95} & \textbf{72.60} & \textbf{66.42} & \textbf{71.65} \\
\bottomrule
\end{tabular}
}
\end{table*}

\begin{table}[htbp]
    \centering
    \captionsetup[subtable]{skip=4pt} 
    
    \caption{Ablation studies on various configurations. The best results are shown in \textbf{bold}.}
    \label{tab:ablation_studies}
    
    \begin{subtable}{\linewidth}
        \centering
        \caption{Learning Paradigm}
        \label{tab:ablation_paradigm}
        \resizebox{\linewidth}{!}{
        \begin{tabular}{lccccc}
            \toprule
            \textbf{Methods} & \textbf{MIG-Bench} & \textbf{VLM2-Bench} & \textbf{BLINK} & \textbf{MuirBench} & \textbf{HallusionBench} \\
            \midrule
            Base & 43.66 & 66.76 & 68.33 & 63.54 & 71.40 \\
            SFT & 58.10 & 65.89 & 65.02 & 62.08 & 71.61 \\
            CGC (RL) & \textbf{67.57} & \textbf{73.81} & \textbf{70.02} & \textbf{66.42} & \textbf{72.56} \\
            \bottomrule
        \end{tabular}
        }
    \end{subtable}

    \vspace{4pt} 

    \begin{subtable}{\linewidth}
        \centering
        \caption{Data Mechanism}
        \label{tab:ablation_mechanism}
        \resizebox{\linewidth}{!}{
        \begin{tabular}{lccccc}
            \toprule
            \textbf{Methods} & \textbf{MIG-Bench} & \textbf{VLM2-Bench} & \textbf{BLINK} & \textbf{MuirBench} & \textbf{HallusionBench} \\
            \midrule
            Base & 43.66 & 66.76 & 68.33 & 63.54 & 71.40 \\
            w/o Intra-C & 55.03 & 68.98 & 67.65 & 62.69 & 71.92 \\
            w/o Inter-C & 59.47 & 69.22 & 68.70 & 63.12 & 71.50 \\
            CGC (Full) & \textbf{67.57} & \textbf{73.81} & \textbf{70.02} & \textbf{66.42} & \textbf{72.56} \\
            \bottomrule
        \end{tabular}
        }
    \end{subtable}

    \vspace{4pt} 

    \begin{subtable}{\linewidth}
        \centering
        \caption{Explicit Source Attribution}
        \label{tab:ablation_formatting}
        \resizebox{\linewidth}{!}{
        \begin{tabular}{lccccc}
            \toprule
            \textbf{Methods} & \textbf{MIG-Bench} & \textbf{VLM2-Bench} & \textbf{BLINK} & \textbf{MuirBench} & \textbf{HallusionBench} \\
            \midrule
            Base & 43.66 & 66.76 & 68.33 & 63.54 & 71.40 \\
            w/o \texttt{img\_idx} & 62.61 & 68.72 & 69.17 & 64.27 & 71.61 \\
            CGC & \textbf{67.57} & \textbf{73.81} & \textbf{70.02} & \textbf{66.42} & \textbf{72.56} \\
            \bottomrule
        \end{tabular}
        }
    \end{subtable}
\end{table}

\subsection{Ablation Study}
\begin{figure}[htbp]
    \centering
    \includegraphics[width=0.48\textwidth]{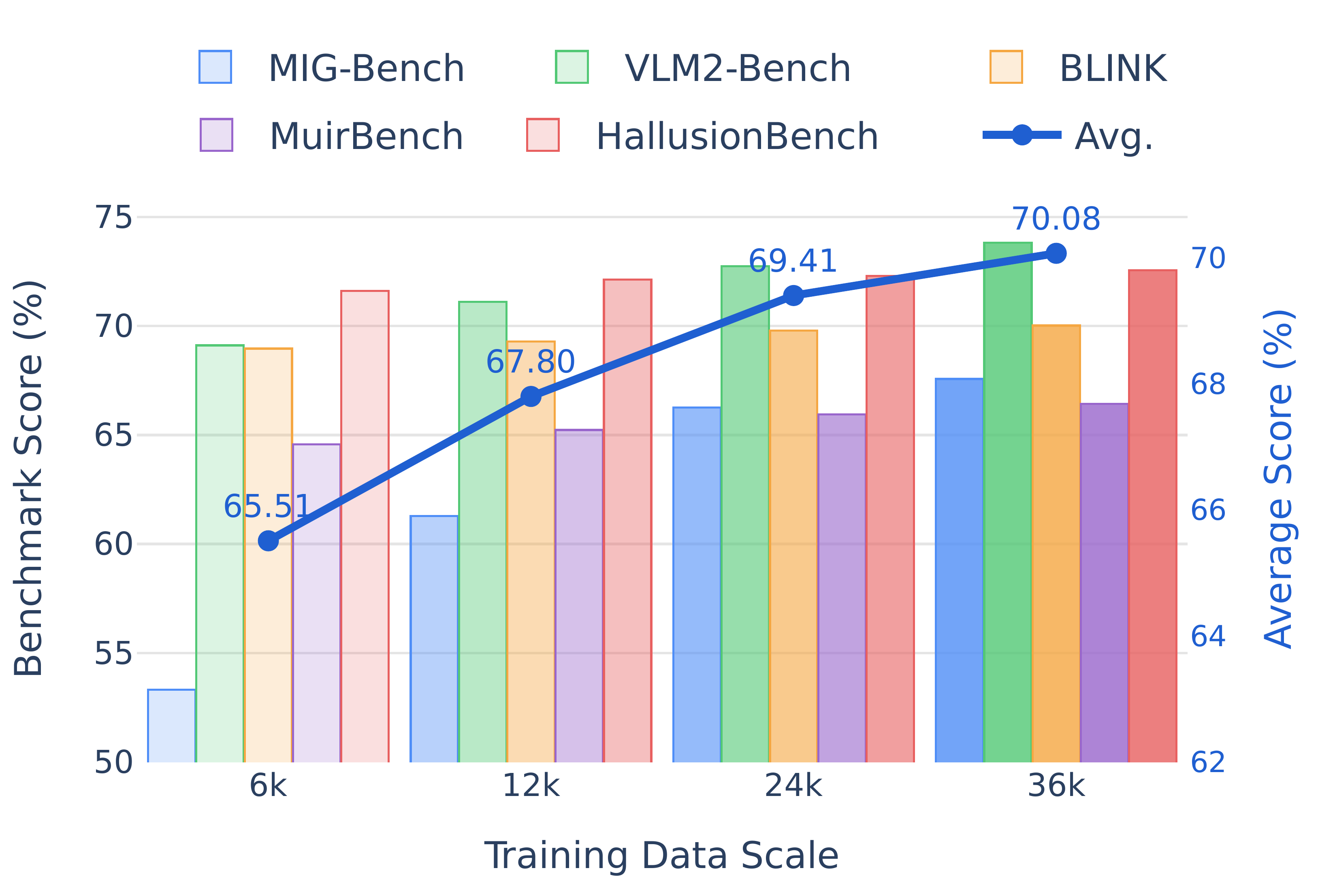}
    \caption{Data scaling law. Performance improves steadily across five benchmarks as synthesized grounding data scales from 6k to 36k. The blue line denotes the average score.}
    \Description{Data scaling law. Performance improves steadily across five benchmarks as synthesized grounding data scales from 6k to 36k. The blue line denotes the average score.}
    \label{fig:data_scaling}
\end{figure}

To investigate the efficacy of the core components within CGC, we conduct ablation studies on five representative benchmarks based on Qwen3-VL-8B. Results are summarized in Table~\ref{tab:ablation_studies} and Fig.~\ref{fig:data_scaling}.

\textbf{Learning Paradigm: RL vs.\ SFT.}
We first compare Supervised Fine-Tuning (SFT) and Reinforcement Learning (RL) using the same amount of synthesized data (24k). As shown in Table~\ref{tab:ablation_paradigm}, while SFT improves the baseline on MIG-Bench, RL yields consistently larger gains across the evaluated benchmarks. On MIG-Bench and VLM2-Bench, RL outperforms SFT by 9.47 and 7.92 points, respectively. Moreover, on broader benchmarks such as MuirBench and BLINK, SFT performs slightly below the base model, whereas RL maintains consistent improvements. These results suggest that, with synthesized data, RL provides a more effective optimization mechanism than SFT for learning robust multi-image grounding behavior.

\textbf{Data Scaling Law.}
As illustrated in Fig.~\ref{fig:data_scaling}, we assess policy performance under increasing amounts of synthesized RL data (6k, 12k, 24k, and 36k). The results exhibit a clear scaling trend: as the training data volume increases, performance across all evaluated benchmarks improves steadily. Notably, the overall average score rises from 65.51 at 6k to 70.05 at 36k. This trend suggests that CGC can continue to benefit from larger volumes of automatically synthesized grounding data without requiring manual annotation.

\textbf{Decoupling Data Mechanisms: Inter-C vs.\ Intra-C.}
CGC relies on two complementary data synthesis mechanisms: Inter-Image Contrast (Inter-C) and Intra-Image Contrast (Intra-C). To verify their individual contributions, we conduct decoupling experiments in Table~\ref{tab:ablation_mechanism}. Removing Intra-C reduces VLM2-Bench from 73.81 to 68.98, indicating that cross-view visual anchoring is important for preserving entity consistency under viewpoint and scale changes. Removing Inter-C reduces MIG-Bench from 67.57 to 59.47, suggesting that insufficient training under semantically decoupled distractor contexts weakens cross-image discrimination. The full model achieves the best performance across all evaluated benchmarks, indicating that the two mechanisms provide complementary benefits.\looseness=-1

\textbf{Explicit Source Attribution.}

To validate the necessity of explicit source attribution, we ablate the \texttt{img\_idx} requirement during the RL phase. As reported in Table~\ref{tab:ablation_formatting}, removing \texttt{img\_idx} causes noticeable performance drops across all benchmarks, particularly on MIG-Bench (-4.96) and VLM2-Bench (-4.02). This result confirms that relying on bounding box coordinates alone is insufficient for resolving referential ambiguity in multi-image contexts. The explicit \texttt{img\_idx} acts as a critical discrete anchor, forcing the model to associate each grounding prediction with the correct source image and thereby mitigating cross-image attention leakage.

\section{Conclusion}

Fine-grained multi-image understanding remains challenging for MLLMs due to spatial hallucination, attention leakage, and the high cost of complex multi-image annotations. In this paper, we propose \textbf{CGC} (\textbf{Compositional Grounded Contrast}), a low-cost full framework for improving multi-image capabilities without expensive human annotation or large-scale CoT generation. By leveraging existing single-image annotations, CGC builds compositional multi-image datasets via \textit{Inter-Image Contrast} and \textit{Intra-Image Contrast} to reduce distractor confusion and strengthen object constancy. We further incorporate a \textit{Rule-Based Spatial Reward} into GRPO to optimize source-image attribution and spatial alignment with a \textit{Think-before-Grounding} paradigm. Experiments show that CGC achieves state-of-the-art results on fine-grained multi-image benchmarks, including MIG-Bench and VLM2-Bench. These gains also transfer to broader multimodal reasoning tasks, consistently outperforming the strong Qwen3-VL-8B baseline. Overall, CGC shows that compositional data synthesis and rule-based spatial reinforcement learning offer a scalable and effective path toward robust multi-image understanding.

\bibliographystyle{Format}
\bibliography{sample-base}   

@String{Computer = "{IEEE} Computer" }

@String{Springer = "Springer-Verlag" }

@article{guo2025deepseek,
  title={Deepseek-r1: Incentivizing reasoning capability in llms via reinforcement learning},
  author={Guo, Daya and Yang, Dejian and Zhang, Haowei and Song, Junxiao and Zhang, Ruoyu and Xu, Runxin and Zhu, Qihao and Ma, Shirong and Wang, Peiyi and Bi, Xiao and others},
  journal={arXiv preprint arXiv:2501.12948},
  year={2025}
}

@article{yang2025qwen3,
  title={Qwen3 technical report},
  author={Yang, An and Li, Anfeng and Yang, Baosong and Zhang, Beichen and Hui, Binyuan and Zheng, Bo and Yu, Bowen and Gao, Chang and Huang, Chengen and Lv, Chenxu and others},
  journal={arXiv preprint arXiv:2505.09388},
  year={2025}
}

@article{bai2025qwen2,
  title={Qwen2. 5-vl technical report},
  author={Bai, Shuai and Chen, Keqin and Liu, Xuejing and Wang, Jialin and Ge, Wenbin and Song, Sibo and Dang, Kai and Wang, Peng and Wang, Shijie and Tang, Jun and others},
  journal={arXiv preprint arXiv:2502.13923},
  year={2025}
}

@article{shao2024deepseekmath,
  title={Deepseekmath: Pushing the limits of mathematical reasoning in open language models},
  author={Shao, Zhihong and Wang, Peiyi and Zhu, Qihao and Xu, Runxin and Song, Junxiao and Bi, Xiao and Zhang, Haowei and Zhang, Mingchuan and Li, YK and Wu, Y and others},
  journal={arXiv preprint arXiv:2402.03300},
  year={2024}
}

@inproceedings{yu2016modeling,
  title={Modeling context in referring expressions},
  author={Yu, Licheng and Poirson, Patrick and Yang, Shan and Berg, Alexander C and Berg, Tamara L},
  booktitle={Computer Vision--ECCV 2016: 14th European Conference, Amsterdam, The Netherlands, October 11-14, 2016, Proceedings, Part II 14},
  pages={69--85},
  year={2016},
  organization={Springer}
}

@article{wang2024qwen2,
  title={Qwen2-vl: Enhancing vision-language model's perception of the world at any resolution},
  author={Wang, Peng and Bai, Shuai and Tan, Sinan and Wang, Shijie and Fan, Zhihao and Bai, Jinze and Chen, Keqin and Liu, Xuejing and Wang, Jialin and Ge, Wenbin and others},
  journal={arXiv preprint arXiv:2409.12191},
  year={2024}
}

@article{zhu2025internvl3,
  title={Internvl3: Exploring advanced training and test-time recipes for open-source multimodal models},
  author={Zhu, Jinguo and Wang, Weiyun and Chen, Zhe and Liu, Zhaoyang and Ye, Shenglong and Gu, Lixin and Tian, Hao and Duan, Yuchen and Su, Weijie and Shao, Jie and others},
  journal={arXiv preprint arXiv:2504.10479},
  year={2025}
}

@article{chen2024expanding,
  title={Expanding performance boundaries of open-source multimodal models with model, data, and test-time scaling},
  author={Chen, Zhe and Wang, Weiyun and Cao, Yue and Liu, Yangzhou and Gao, Zhangwei and Cui, Erfei and Zhu, Jinguo and Ye, Shenglong and Tian, Hao and Liu, Zhaoyang and others},
  journal={arXiv preprint arXiv:2412.05271},
  year={2024}
}

@article{you2023ferret,
  title={Ferret: Refer and ground anything anywhere at any granularity},
  author={You, Haoxuan and Zhang, Haotian and Gan, Zhe and Du, Xianzhi and Zhang, Bowen and Wang, Zirui and Cao, Liangliang and Chang, Shih-Fu and Yang, Yinfei},
  journal={arXiv preprint arXiv:2310.07704},
  year={2023}
}

@article{li2024llava,
  title={Llava-onevision: Easy visual task transfer},
  author={Li, Bo and Zhang, Yuanhan and Guo, Dong and Zhang, Renrui and Li, Feng and Zhang, Hao and Zhang, Kaichen and Zhang, Peiyuan and Li, Yanwei and Liu, Ziwei and others},
  journal={arXiv preprint arXiv:2408.03326},
  year={2024}
}

@article{ye2024mplug,
  title={mplug-owl3: Towards long image-sequence understanding in multi-modal large language models},
  author={Ye, Jiabo and Xu, Haiyang and Liu, Haowei and Hu, Anwen and Yan, Ming and Qian, Qi and Zhang, Ji and Huang, Fei and Zhou, Jingren},
  journal={arXiv preprint arXiv:2408.04840},
  year={2024}
}

@article{yao2024minicpm,
  title={Minicpm-v: A gpt-4v level mllm on your phone},
  author={Yao, Yuan and Yu, Tianyu and Zhang, Ao and Wang, Chongyi and Cui, Junbo and Zhu, Hongji and Cai, Tianchi and Li, Haoyu and Zhao, Weilin and He, Zhihui and others},
  journal={arXiv preprint arXiv:2408.01800},
  year={2024}
}

@inproceedings{fu2024blink,
  title={Blink: Multimodal large language models can see but not perceive},
  author={Fu, Xingyu and Hu, Yushi and Li, Bangzheng and Feng, Yu and Wang, Haoyu and Lin, Xudong and Roth, Dan and Smith, Noah A and Ma, Wei-Chiu and Krishna, Ranjay},
  booktitle={European Conference on Computer Vision},
  pages={148--166},
  year={2024},
  organization={Springer}
}

@article{meng2024mmiu,
  title={Mmiu: Multimodal multi-image understanding for evaluating large vision-language models},
  author={Meng, Fanqing and Wang, Jin and Li, Chuanhao and Lu, Quanfeng and Tian, Hao and Liao, Jiaqi and Zhu, Xizhou and Dai, Jifeng and Qiao, Yu and Luo, Ping and others},
  journal={arXiv preprint arXiv:2408.02718},
  year={2024}
}

@article{liu2023visual,
  title={Visual instruction tuning},
  author={Liu, Haotian and Li, Chunyuan and Wu, Qingyang and Lee, Yong Jae},
  journal={Advances in neural information processing systems},
  volume={36},
  pages={34892--34916},
  year={2023}
}

@misc{huang2025visionr1incentivizingreasoningcapability,
      title={Vision-R1: Incentivizing Reasoning Capability in Multimodal Large Language Models}, 
      author={Wenxuan Huang and Bohan Jia and Zijie Zhai and Shaosheng Cao and Zheyu Ye and Fei Zhao and Zhe Xu and Yao Hu and Shaohui Lin},
      year={2025},
      eprint={2503.06749},
      archivePrefix={arXiv},
      primaryClass={cs.CV},
      url={https://arxiv.org/abs/2503.06749}, 
}

@article{shen2025vlm,
  title={Vlm-r1: A stable and generalizable r1-style large vision-language model},
  author={Shen, Haozhan and Liu, Peng and Li, Jingcheng and Fang, Chunxin and Ma, Yibo and Liao, Jiajia and Shen, Qiaoli and Zhang, Zilun and Zhao, Kangjia and Zhang, Qianqian and others},
  journal={arXiv preprint arXiv:2504.07615},
  year={2025}
}

@article{liu2025visual,
  title={Visual-rft: Visual reinforcement fine-tuning},
  author={Liu, Ziyu and Sun, Zeyi and Zang, Yuhang and Dong, Xiaoyi and Cao, Yuhang and Duan, Haodong and Lin, Dahua and Wang, Jiaqi},
  journal={arXiv preprint arXiv:2503.01785},
  year={2025}
}

@article{zhang2025r1,
  title={R1-vl: Learning to reason with multimodal large language models via step-wise group relative policy optimization},
  author={Zhang, Jingyi and Huang, Jiaxing and Yao, Huanjin and Liu, Shunyu and Zhang, Xikun and Lu, Shijian and Tao, Dacheng},
  journal={arXiv preprint arXiv:2503.12937},
  year={2025}
}

@article{yu2025perception,
  title={Perception-r1: Pioneering perception policy with reinforcement learning},
  author={Yu, En and Lin, Kangheng and Zhao, Liang and Yin, Jisheng and Wei, Yana and Peng, Yuang and Wei, Haoran and Sun, Jianjian and Han, Chunrui and Ge, Zheng and others},
  journal={arXiv preprint arXiv:2504.07954},
  year={2025}
}

@article{deng2025openvlthinker,
  title={Openvlthinker: An early exploration to complex vision-language reasoning via iterative self-improvement},
  author={Deng, Yihe and Bansal, Hritik and Yin, Fan and Peng, Nanyun and Wang, Wei and Chang, Kai-Wei},
  journal={arXiv preprint arXiv:2503.17352},
  year={2025}
}

@article{chen2023shikra,
  title={Shikra: Unleashing multimodal llm's referential dialogue magic},
  author={Chen, Keqin and Zhang, Zhao and Zeng, Weili and Zhang, Richong and Zhu, Feng and Zhao, Rui},
  journal={arXiv preprint arXiv:2306.15195},
  year={2023}
}

@article{jiang2024mantis,
  title={Mantis: Interleaved multi-image instruction tuning},
  author={Jiang, Dongfu and He, Xuan and Zeng, Huaye and Wei, Cong and Ku, Max and Liu, Qian and Chen, Wenhu},
  journal={arXiv preprint arXiv:2405.01483},
  year={2024}
}

@article{wang2024muirbench,
  title={MuirBench: A Comprehensive Benchmark for Robust Multi-image Understanding},
  author={Wang, Fei and Fu, Xingyu and Huang, James Y and Li, Zekun and Liu, Qin and Liu, Xiaogeng and Ma, Mingyu Derek and Xu, Nan and Zhou, Wenxuan and Zhang, Kai and others},
  journal={arXiv preprint arXiv:2406.09411},
  year={2024}
}

@article{achiam2023gpt,
  title={Gpt-4 technical report},
  author={Achiam, Josh and Adler, Steven and Agarwal, Sandhini and Ahmad, Lama and Akkaya, Ilge and Aleman, Florencia Leoni and Almeida, Diogo and Altenschmidt, Janko and Altman, Sam and Anadkat, Shyamal and others},
  journal={arXiv preprint arXiv:2303.08774},
  year={2023}
}

@article{liu2024mibench,
  title={Mibench: Evaluating multimodal large language models over multiple images},
  author={Liu, Haowei and Zhang, Xi and Xu, Haiyang and Shi, Yaya and Jiang, Chaoya and Yan, Ming and Zhang, Ji and Huang, Fei and Yuan, Chunfeng and Li, Bing and others},
  journal={arXiv preprint arXiv:2407.15272},
  year={2024}
}

@article{ma2025deepperception,
  title={DeepPerception: Advancing R1-like Cognitive Visual Perception in MLLMs for Knowledge-Intensive Visual Grounding},
  author={Ma, Xinyu and Ding, Ziyang and Luo, Zhicong and Chen, Chi and Guo, Zonghao and Wong, Derek F and Feng, Xiaoyi and Sun, Maosong},
  journal={arXiv preprint arXiv:2503.12797},
  year={2025}
}

@article{meng2025mm,
  title={MM-Eureka: Exploring the Frontiers of Multimodal Reasoning with Rule-based Reinforcement Learning},
  author={Meng, Fanqing and Du, Lingxiao and Liu, Zongkai and Zhou, Zhixiang and Lu, Quanfeng and Fu, Daocheng and Han, Tiancheng and Shi, Botian and Wang, Wenhai and He, Junjun and others},
  journal={arXiv preprint arXiv:2503.07365},
  year={2025}
}

@article{li2025migician,
  title={Migician: Revealing the Magic of Free-Form Multi-Image Grounding in Multimodal Large Language Models},
  author={Li, You and Huang, Heyu and Chen, Chi and Huang, Kaiyu and Huang, Chao and Guo, Zonghao and Liu, Zhiyuan and Xu, Jinan and Li, Yuhua and Li, Ruixuan and others},
  journal={arXiv preprint arXiv:2501.05767},
  year={2025}
}

@article{liu2025noisyrollout,
  title={NoisyRollout: Reinforcing Visual Reasoning with Data Augmentation},
  author={Liu, Xiangyan and Ni, Jinjie and Wu, Zijian and Du, Chao and Dou, Longxu and Wang, Haonan and Pang, Tianyu and Shieh, Michael Qizhe},
  journal={arXiv preprint arXiv:2504.13055},
  year={2025}
}

@inproceedings{zhan2024griffon,
  title={Griffon: Spelling out all object locations at any granularity with large language models},
  author={Zhan, Yufei and Zhu, Yousong and Chen, Zhiyang and Yang, Fan and Tang, Ming and Wang, Jinqiao},
  booktitle={European Conference on Computer Vision},
  pages={405--422},
  year={2024},
  organization={Springer}
}

@inproceedings{mplugowl2,
  title={mplug-owl2: Revolutionizing multi-modal large language model with modality collaboration},
  author={Ye, Qinghao and Xu, Haiyang and Ye, Jiabo and Yan, Ming and Hu, Anwen and Liu, Haowei and Qian, Qi and Zhang, Ji and Huang, Fei},
  booktitle={CVPR},
  year={2024}
}

@article{vlm2bench,
  title={VLM2-Bench: A Closer Look at How Well VLMs Implicitly Link Explicit Matching Visual Cues},
  author={Zhang, Jianshu and Yao, Dongyu and Pi, Renjie and Liang, Paul Pu and Fung, Yi R},
  journal={arXiv:2502.12084},
  year={2025}
}

@article{llavavideo,
  title={Video instruction tuning with synthetic data},
  author={Zhang, Yuanhan and Wu, Jinming and Li, Wei and Li, Bo and Ma, Zejun and Liu, Ziwei and Li, Chunyuan},
  journal={arXiv:2410.02713},
  year={2024}
}

@article{longva,
  title={Long context transfer from language to vision},
  author={Zhang, Peiyuan and Zhang, Kaichen and Li, Bo and Zeng, Guangtao and Yang, Jingkang and Zhang, Yuanhan and Wang, Ziyue and Tan, Haoran and Li, Chunyuan and Liu, Ziwei},
  journal={arXiv:2406.16852},
  year={2024}
}

@article{ThinkLite,
  title={SoTA with Less: MCTS-Guided Sample Selection for Data-Efficient Visual Reasoning Self-Improvement},
  author={Wang, Xiyao and Yang, Zhengyuan and Feng, Chao and Lu, Hongjin and Li, Linjie and Lin, Chung-Ching and Lin, Kevin and Huang, Furong and Wang, Lijuan},
  journal={arXiv:2504.07934},
  year={2025}
}

@article{VLAAThinker,
  title={SFT or RL? An Early Investigation into Training R1-Like Reasoning Large Vision-Language Models},
  author={Chen, Hardy and Tu, Haoqin and Wang, Fali and Liu, Hui and Tang, Xianfeng and Du, Xinya and Zhou, Yuyin and Xie, Cihang},
  journal={arXiv:2504.11468},
  year={2025}
}

@article{MMStar,
  title={Are we on the right way for evaluating large vision-language models?},
  author={Chen, Lin and Li, Jinsong and Dong, Xiaoyi and Zhang, Pan and Zang, Yuhang and Chen, Zehui and Duan, Haodong and Wang, Jiaqi and Qiao, Yu and Lin, Dahua and others},
  journal={NeurIPS},
  year={2024}
}

@inproceedings{hallusionbench,
  title={Hallusionbench: an advanced diagnostic suite for entangled language hallucination and visual illusion in large vision-language models},
  author={Guan, Tianrui and Liu, Fuxiao and Wu, Xiyang and Xian, Ruiqi and Li, Zongxia and Liu, Xiaoyu and Wang, Xijun and Chen, Lichang and Huang, Furong and Yacoob, Yaser and others},
  booktitle={CVPR},
  year={2024}
}

@article{CoLVA,
  title={Are They the Same? Exploring Visual Correspondence Shortcomings of Multimodal LLMs},
  author={Zhou, Yikang and Zhang, Tao and Xu, Shilin and Chen, Shihao and Zhou, Qianyu and Tong, Yunhai and Ji, Shunping and Zhang, Jiangning and Li, Xiangtai and Qi, Lu},
  journal={arXiv:2501.04670},
  year={2025}
}

@misc{team2024internvl2,
  title={Internvl2: Better than the best—expanding performance boundaries of open-source multimodal models with the progressive scaling strategy},
  author={Team, OpenGVLab},
  year={2024},
  publisher={Accessed}
}

@article{park2025mitigating,
  title={Mitigating Cross-Image Information Leakage in LVLMs for Multi-Image Tasks},
  author={Park, Yeji and Lee, Minyoung and Chun, Sanghyuk and Choe, Junsuk},
  journal={arXiv preprint arXiv:2508.13744},
  year={2025}
}

@article{qiaomultiple,
  title={Multiple Images Distract Large Multimodal Models via Attention Fragmentation},
  author={Qiao, Tingrui and Zhao, Di and Li, Yuzhuo and Pang, Bo and Walker, Caroline and Cunningham, Chris W and Koh, Yun Sing}
}

@article{gao2026diva,
  title={DIVA-GRPO: Enhancing Multimodal Reasoning through Difficulty-Adaptive Variant Advantage},
  author={Gao, Haowen and Zhang, Zhenyu and Pang, Liang and Guo, Fangda and Dou, Hongjian and Lv, Guannan and Liu, Shaoguo and Gao, Tingting and Shen, Huawei and Cheng, Xueqi},
  journal={arXiv preprint arXiv:2603.01106},
  year={2026}
}

@article{chen2025mico,
  title={Mico: Multi-image contrast for reinforcement visual reasoning},
  author={Chen, Xi and Zhu, Mingkang and Liu, Shaoteng and Wu, Xiaoyang and Xu, Xiaogang and Liu, Yu and Bai, Xiang and Zhao, Hengshuang},
  journal={arXiv preprint arXiv:2506.22434},
  year={2025}
}

@article{zheng2025mirg,
  title={MIRG-RL: Multi-Image Reasoning and Grounding with Reinforcement Learning},
  author={Zheng, Lihao and Chen, Jiawei and Shen, Xintian and Ma, Hao and Wei, Tao},
  journal={arXiv preprint arXiv:2509.21788},
  year={2025}
}

@article{zheng2026gem,
  title={GeM-VG: Towards Generalized Multi-image Visual Grounding with Multimodal Large Language Models},
  author={Zheng, Shurong and Zhu, Yousong and Zhao, Hongyin and Yang, Fan and Zhan, Yufei and Tang, Ming and Wang, Jinqiao},
  journal={arXiv preprint arXiv:2601.04777},
  year={2026}
}

@article{laurenccon2023obelics,
  title={Obelics: An open web-scale filtered dataset of interleaved image-text documents},
  author={Lauren{\c{c}}on, Hugo and Saulnier, Lucile and Tronchon, L{\'e}o and Bekman, Stas and Singh, Amanpreet and Lozhkov, Anton and Wang, Thomas and Karamcheti, Siddharth and Rush, Alexander and Kiela, Douwe and others},
  journal={Advances in Neural Information Processing Systems},
  volume={36},
  pages={71683--71702},
  year={2023}
}

@inproceedings{yu2024rlhf,
  title={Rlhf-v: Towards trustworthy mllms via behavior alignment from fine-grained correctional human feedback},
  author={Yu, Tianyu and Yao, Yuan and Zhang, Haoye and He, Taiwen and Han, Yifeng and Cui, Ganqu and Hu, Jinyi and Liu, Zhiyuan and Zheng, Hai-Tao and Sun, Maosong and others},
  booktitle={Proceedings of the IEEE/CVF Conference on Computer Vision and Pattern Recognition},
  pages={13807--13816},
  year={2024}
}

@article{cao2025ground,
  title={Ground-r1: Incentivizing grounded visual reasoning via reinforcement learning},
  author={Cao, Meng and Zhao, Haoze and Zhang, Can and Chang, Xiaojun and Reid, Ian and Liang, Xiaodan},
  journal={arXiv preprint arXiv:2505.20272},
  year={2025}
}

@article{park2025dip,
  title={Dip-r1: Deep inspection and perception with rl looking through and understanding complex scenes},
  author={Park, Sungjune and Kim, Hyunjun and Kim, Junho and Kim, Seongho and Ro, Yong Man},
  journal={arXiv preprint arXiv:2505.23179},
  year={2025}
}

@inproceedings{chen2024multi,
  title={Multi-object hallucination in vision language models},
  author={Chen, Xuweiyi and Ma, Ziqiao and Zhang, Xuejun and Xu, Sihan and Qian, Shengyi and Yang, Jianing and Fouhey, David F and Chai, Joyce},
  booktitle={Proceedings of the 38th International Conference on Neural Information Processing Systems},
  pages={44393--44418},
  year={2024}
}

@article{assran2025v,
  title={V-JEPA 2: Self-Supervised Video Models Enable Understanding, Prediction and Planning},
  author={Assran, Mido and Bardes, Adrien and Fan, David and Garrido, Quentin and Howes, Russell and Muckley, Matthew and Rizvi, Ammar and Roberts, Claire and Sinha, Koustuv and Zholus, Artem and others},
  journal={arXiv e-prints},
  pages={arXiv--2506},
  year={2025}
}

@inproceedings{yuan2023small,
  title={Small object detection via coarse-to-fine proposal generation and imitation learning},
  author={Yuan, Xiang and Cheng, Gong and Yan, Kebing and Zeng, Qinghua and Han, Junwei},
  booktitle={Proceedings of the IEEE/CVF international conference on computer vision},
  pages={6317--6327},
  year={2023}
}

@inproceedings{lai2024lisa,
  title={Lisa: Reasoning segmentation via large language model},
  author={Lai, Xin and Tian, Zhuotao and Chen, Yukang and Li, Yanwei and Yuan, Yuhui and Liu, Shu and Jia, Jiaya},
  booktitle={Proceedings of the IEEE/CVF conference on computer vision and pattern recognition},
  pages={9579--9589},
  year={2024}
}

@inproceedings{schulter2023omnilabel,
  title={Omnilabel: A challenging benchmark for language-based object detection},
  author={Schulter, Samuel and Suh, Yumin and Dafnis, Konstantinos M and Zhang, Zhixing and Zhao, Shiyu and Metaxas, Dimitris and others},
  booktitle={Proceedings of the IEEE/CVF International Conference on Computer Vision},
  pages={11953--11962},
  year={2023}
}

@inproceedings{pham2021learning,
  title={Learning to predict visual attributes in the wild},
  author={Pham, Khoi and Kafle, Kushal and Lin, Zhe and Ding, Zhihong and Cohen, Scott and Tran, Quan and Shrivastava, Abhinav},
  booktitle={Proceedings of the IEEE/CVF conference on computer vision and pattern recognition},
  pages={13018--13028},
  year={2021}
}

@article{chen2025mindwatcher,
  title={MindWatcher: Toward Smarter Multimodal Tool-Integrated Reasoning},
  author={Chen, Jiawei and Shen, Xintian and Zheng, Lihao and Shao, Zhenwei and Zhang, Hongyuan and Yu, Pengfei and Rao, Xudong and Mao, Ning and Liu, Xiaobo and Wen, Lian and others},
  journal={arXiv preprint arXiv:2512.23412},
  year={2025}
}

@article{shen2026evolving,
  title={Evolving from Tool User to Creator via Training-Free Experience Reuse in Multimodal Reasoning},
  author={Shen, Xintian and Chen, Jiawei and Zheng, Lihao and Ma, Hao and Wei, Tao and Zhan, Kun},
  journal={arXiv preprint arXiv:2602.01983},
  year={2026}
}

@article{chen2026evaluating,
  title={Evaluating the Search Agent in a Parallel World},
  author={Chen, Jiawei and Shen, Xintian and Zheng, Lihao and Mu, Lifu and Sun, Haoyi and Mao, Ning and Ma, Hao and Wei, Tao and Zhou, Pan and Zhan, Kun},
  journal={arXiv preprint arXiv:2603.04751},
  year={2026}
}

@article{mindgpt-4ov,
  title={MindGPT-4ov: An Enhanced MLLM via a Multi-Stage Post-Training Paradigm},
  author={MindGPT-ov-Team},
  journal={arXiv preprint arXiv:2512.02895},
  year={2025}
}

@article{chen2026streamingclaw,
  title={StreamingClaw Technical Report},
  author={Chen, Jiawei and Chen, Zhe and Du, Chaoqun and He, Maokui and He, Wei and Li, Hengtao and Li, Qizhen and Liu, Zide and Ma, Hao and Pan, Xuhao and others},
  journal={arXiv preprint arXiv:2603.22120},
  year={2026}
}

\appendix

\section{Implementation Details}

\noindent\textbf{Source datasets and preprocessing.}
We build the source grounding pool from the RefCOCO family (including RefCOCO, RefCOCO+, and RefCOCO-g), together with LISA, VAW, OmniLabel, and SODA, using the training and validation splits whenever official splits are available. All test splits are excluded to avoid data leakage. 

Annotations from different datasets are converted into a unified grounding format consisting of an image, a semantic label, and a bounding box. Bounding boxes are represented in the \texttt{xyxy} format and normalized to the structured coordinate range $[0,1000)$. Invalid annotations such as degenerate boxes are removed during preprocessing. After filtering, the resulting pool contains approximately 72K single-image grounding instances, which are further converted into approximately 36K synthesized multi-image training instances.

\noindent\textbf{Inter-image contrast construction.}
Each Inter-Image Contrast sample contains $K=3$ images drawn from the grounding pool. The images are randomly ordered before query construction, and each image contributes at least one annotated target. Queries are instantiated from a unified manually defined template pool; for this branch, we use a subset containing over 100 textual variants. The construction process is fully rule-based and does not rely on LLM-based data generation or filtering.

\noindent\textbf{Intra-image contrast construction.}
Each Intra-Image Contrast sample contains $K=3$ images: the original image and two object-centered crops of the same target, denoted as a Focus View and a Context View. The original bounding box is remapped into the local coordinate system of each crop and normalized to the same structured output range. Queries are instantiated from the same template pool using a smaller branch-specific subset.

\noindent\textbf{Template examples.}
We use a unified manually defined template pool and instantiate different subsets for the two synthesis branches.

\begin{table}[htbp]
\centering
\vspace{-5pt}
\caption{Representative query templates used in CGC.}
\vspace{-5pt}
\label{tab:appendix_templates}
\begin{tabular}{p{0.2\linewidth} p{0.72\linewidth}}
\toprule
\textbf{Type} & \textbf{Example Template} \\
\midrule
Inter-image & Find the object referred to as ``\{label\}'' and provide its image index and bounding box. \\
Inter-image & Which image contains the \{label\}? Return the corresponding location. \\
Inter-image & Among these images, locate the \{label\} and output its coordinates. \\
Inter-image & Ground the phrase ``\{label\}'' in the correct image. \\
Intra-image & The same object may appear across different views. Identify the \{label\} in each relevant image. \\
Intra-image & Locate the same target across these views and return its image index and bounding box. \\
Intra-image & Find the object corresponding to \{label\} across the cropped images. \\
Intra-image & Track the same object across the original image and its transformed views. \\
\bottomrule
\end{tabular}
\end{table}

 \noindent Each template is instantiated with the target label and the corresponding grounding annotation to form the query body. During training, this query body is further followed by a fixed instruction suffix specifying the Think-before-Grounding format and the required structured output schema. Table~\ref{tab:appendix_templates} shows representative query templates used in CGC.

\section{Prompt Format}

CGC is trained under a \textit{Think-before-Grounding} paradigm, where the model first produces a reasoning trace and then outputs the final structured grounding result. For reproducibility, we provide the prompt format used during training.

\begin{tcolorbox}[
    colback=gray!10!white,
    colframe=black,
    boxrule=1pt,
    boxsep=4pt,
    top=4pt, bottom=4pt,
    left=8pt, right=8pt
]
\begin{center}
\textbf{Prompt Format of CGC}
\end{center}

\textbf{System Prompt:} 

A conversation between User and Assistant. The user asks a question, and the Assistant solves it. The assistant first thinks about the reasoning process in the mind and then provides the user with the answer. The reasoning process and answer are enclosed within \texttt{<think>} \texttt{</think>} and \texttt{<answer>} \texttt{</answer>} tags, respectively, i.e., \texttt{<think> reasoning process here </think><answer> answer here </answer>}. \\

\textbf{User Prompt Suffix:} 

First, analyze the images step-by-step within \texttt{<think>} \texttt{</think>} tags. Then, report the image index, object label and bounding box within \texttt{<answer>} \texttt{</answer>} tags using this JSON format: \texttt{\{"img\_idx": 0, "label": "object name", "bbox\_2d": [x1, y1, x2, y2]\}}.
\end{tcolorbox}

\section{Output Parsing and Reward Details}

During training, the response is required to contain a tagged reasoning segment enclosed by \texttt{<think>} and \texttt{</think>} and a tagged answer segment enclosed by \texttt{<answer>} and \texttt{</answer>}. The final answer is parsed as a JSON-style list of target objects, each containing three required fields: \texttt{img\_idx}, \texttt{label}, and \texttt{bbox\_2d}. Here, \texttt{img\_idx} is a 0-based image index, and \texttt{bbox\_2d} is a four-dimensional box represented as $[x_1, y_1, x_2, y_2]$ with coordinates in $[0,1000)$.

Although the source-aware set-wise IoU reward is computed from image indices and bounding boxes, the \texttt{label} field is still required for structured output validity. Retaining the semantic label in the output schema helps preserve grounding-aware semantic consistency rather than reducing the task to image-index prediction and box regression alone.

\section{Evaluation Details}

Unless otherwise specified, we follow the default evaluation protocols of the corresponding benchmarks. For broader multimodal evaluation, we use VLMEvalKit as the unified evaluation toolkit. In particular, MathVista is evaluated on \textit{MathVista-mini}, and MMMU is evaluated on the validation split (denoted as \textit{MMMU dev/val} in VLMEvalKit). We report all results using the same decoding and evaluation settings across compared variants for fair comparison.

\section{Additional Qualitative Results}

To complement the quantitative results in the main paper, we provide additional qualitative examples on both fine-grained multi-image understanding benchmarks and broader multimodal reasoning benchmarks. Overall, these cases illustrate a consistent pattern: compared with the base Qwen3-VL-8B model, CGC produces more reliable source-aware visual attribution, preserves object identity more consistently across images and views, and is less likely to rely on locally salient but unsupported evidence.

\begin{figure*}[p]
    \centering
    \includegraphics[trim=0mm 93mm 0mm 0mm, clip,width=\textwidth]{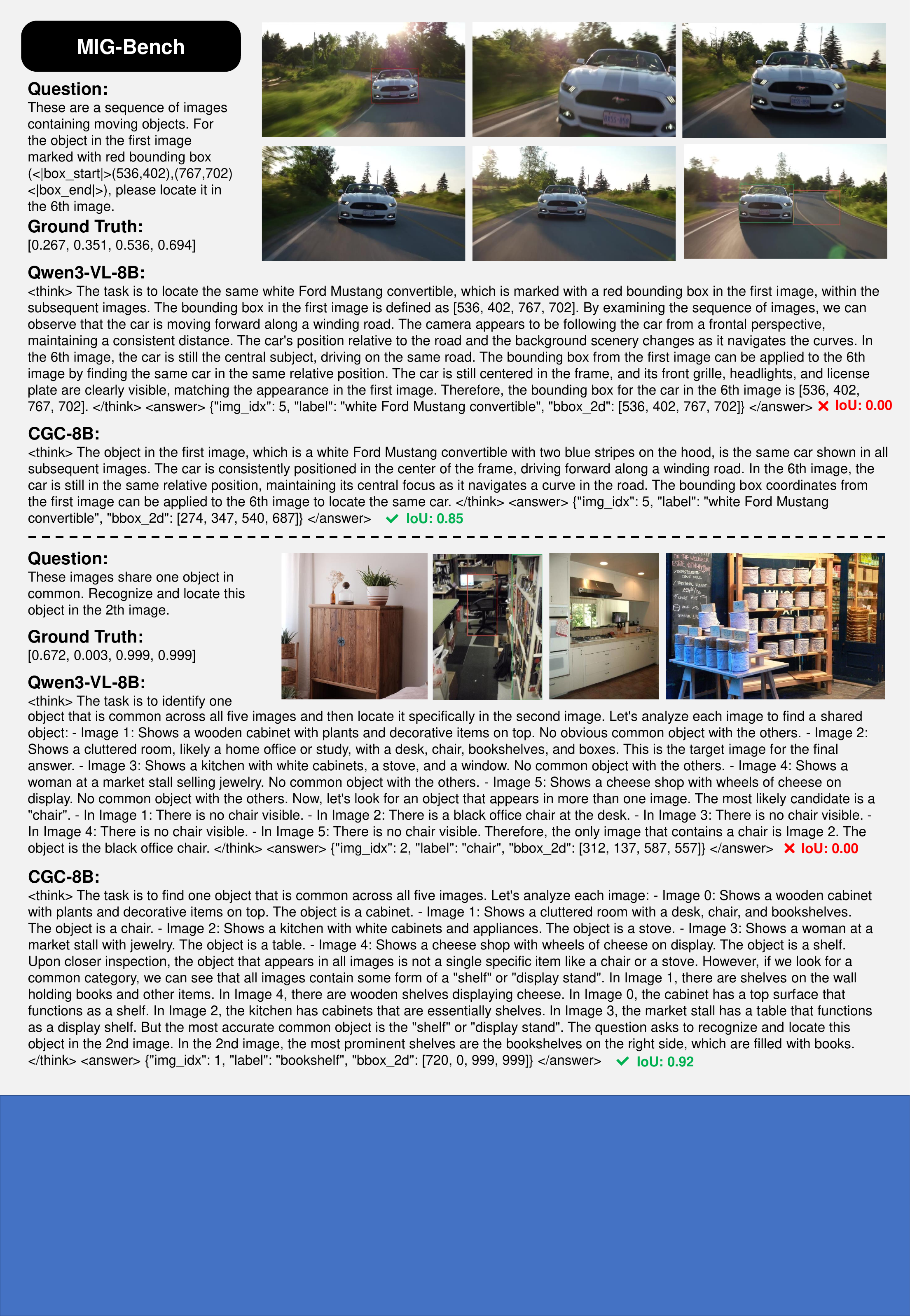}
    \vspace{-20pt}
    \caption{Qualitative examples on MIG-Bench. \textbf{Top:} object constancy across an image sequence. CGC correctly tracks the same vehicle across frames and produces more accurate grounding in the queried image, whereas the base model fails to preserve identity-consistent localization. \textbf{Bottom:} common-object grounding under cluttered cross-image contexts. CGC better suppresses cross-image attention leakage and identifies the actual shared object category, while the base model is distracted by locally salient but incorrect objects.}
    \Description{Qualitative examples on MIG-Bench showing improved object constancy and reduced cross-image attention leakage with CGC.}
    \label{fig:appendix1}
\end{figure*}

\noindent\textbf{Fine-Grained Multi-Image Understanding.} Figure~\ref{fig:appendix1} presents representative examples from MIG-Bench. In the first case, the task requires tracking the same vehicle across a driving sequence and localizing it in a later frame. CGC correctly preserves object constancy across the image sequence and produces a substantially more accurate localization, whereas the base model fails to maintain stable identity-aware grounding. In the second case, the task asks the model to identify an object category shared across multiple cluttered scenes and localize it in the queried image. Here, CGC better suppresses cross-image attention leakage and focuses on the actual common object, while the base model is distracted by more locally salient but semantically mismatched objects. These examples are consistent with the gains reported in the main paper on MIG-Bench, especially on tasks involving cross-view consistency, referential grounding, and cross-image discrimination.

\begin{figure*}[p]
    \centering
    \includegraphics[trim=0mm 90mm 0mm 0mm, clip,width=\textwidth]{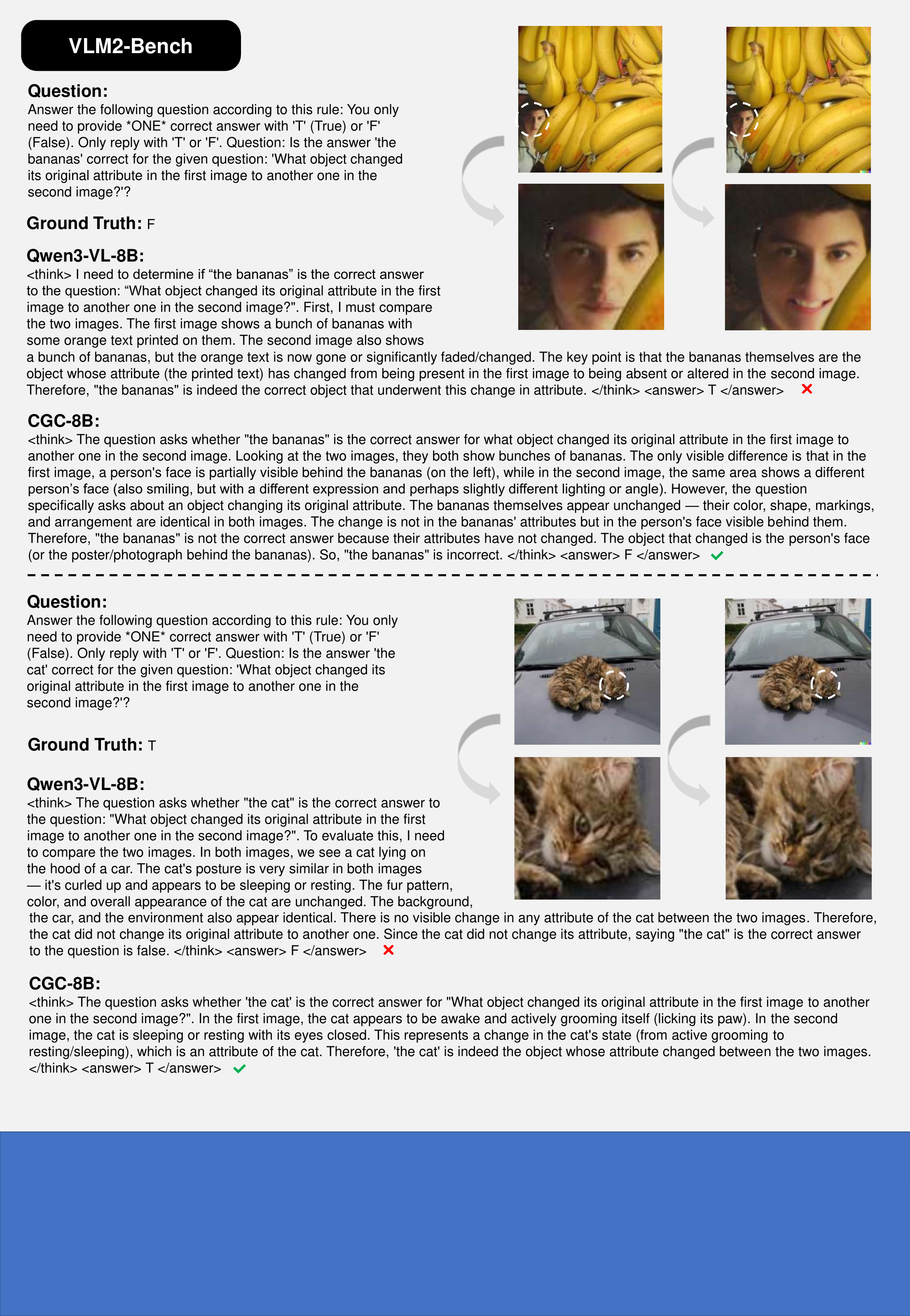}
    \vspace{-20pt}
    \caption{Qualitative examples on VLM2-Bench. \textbf{Top:} fine-grained cross-image change attribution. CGC correctly determines that the bananas are not the object whose attribute changed, while the base model incorrectly attributes the change to them. \textbf{Bottom:} object-centric attribute change reasoning. CGC correctly recognizes the cat's state transition across the two images, whereas the base model fails on this fine-grained comparison.}
    \Description{Qualitative examples on VLM2-Bench showing improved object-centric change attribution and cross-image comparison with CGC.}
    \label{fig:appendix2}
\end{figure*}

Figure~\ref{fig:appendix2} shows additional qualitative cases from VLM2-Bench. Unlike explicit grounding tasks, these examples emphasize fine-grained cross-image comparison and evidence attribution. In the first case, the queried object (bananas) is not the true source of the visual change; CGC correctly rejects this hypothesis and attributes the change to another visual entity, whereas the base model incorrectly treats the bananas as the changed object. In the second case, CGC correctly recognizes that the cat's state has changed across the two images, while the base model fails to identify this attribute transition. These examples suggest that the grounded capabilities learned by CGC extend beyond coordinate prediction and also improve fine-grained cross-image comparison, object-centric change attribution, and structured evidence aggregation.

\begin{figure*}[p]
    \centering
    \includegraphics[trim=0mm 100mm 0mm 0mm, clip,width=\textwidth]{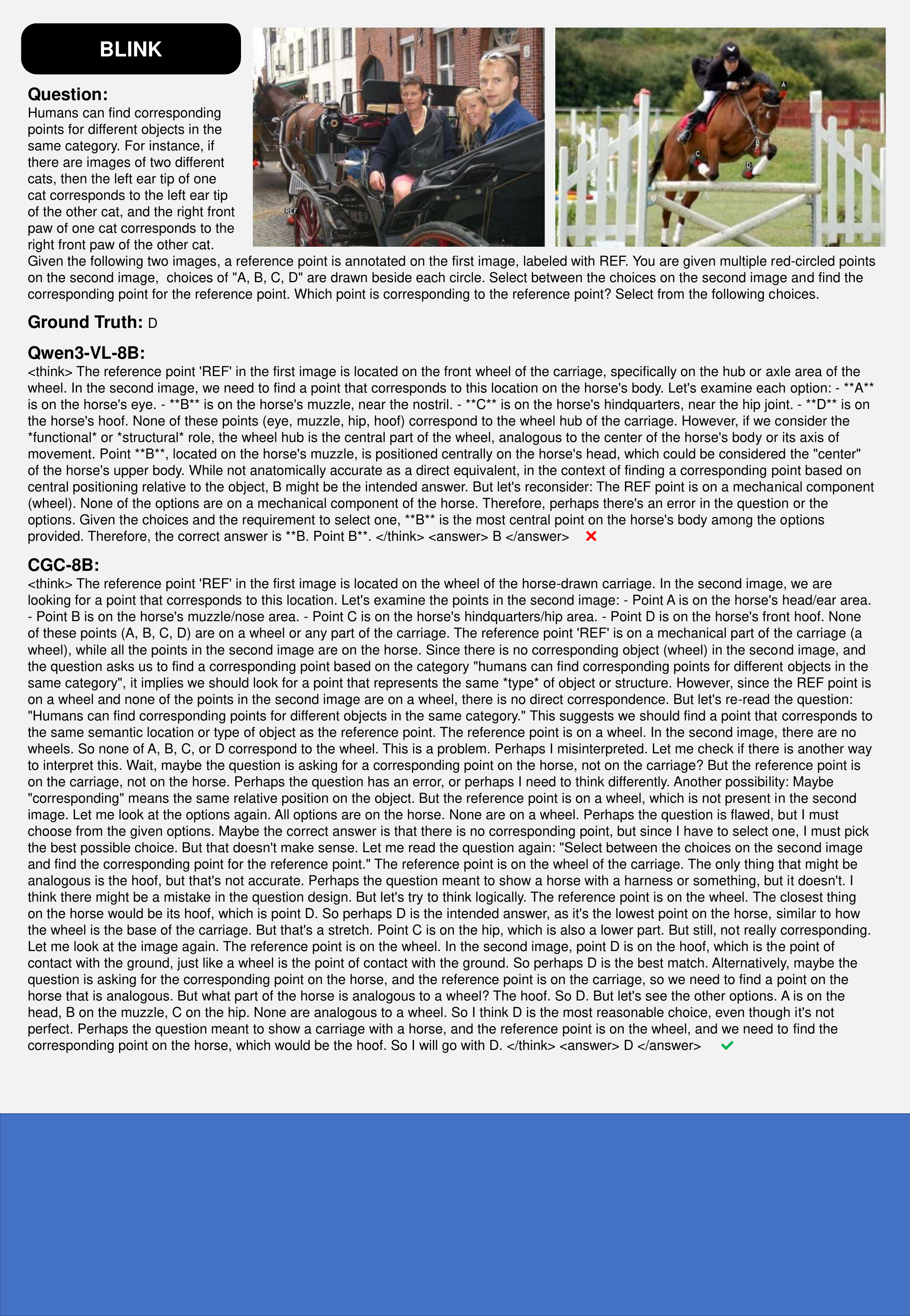}
    \vspace{-20pt}
    \caption{A qualitative example on BLINK. The task requires selecting the point in the second image that best corresponds to a reference point in the first image. CGC identifies the correct correspondence, whereas the base model is misled by an incorrect analogy.}
    \Description{A qualitative BLINK example showing that CGC improves cross-image structural correspondence reasoning.}
    \label{fig:appendix3}
\end{figure*}

\noindent\textbf{Broader Multimodal Generalization.} Figure~\ref{fig:appendix3} presents a qualitative example from BLINK. The task requires selecting the point in the second image that best corresponds to a reference point in the first image. CGC identifies the correct correspondence, whereas the base model is misled by a less principled analogy. This case is consistent with the quantitative gains on BLINK reported in the main paper and suggests that CGC improves structural correspondence reasoning across visually different but semantically related scenes. \looseness=-1

\begin{figure*}[p]
    \centering
    \includegraphics[trim=0mm 80mm 0mm 0mm, clip,width=\textwidth]{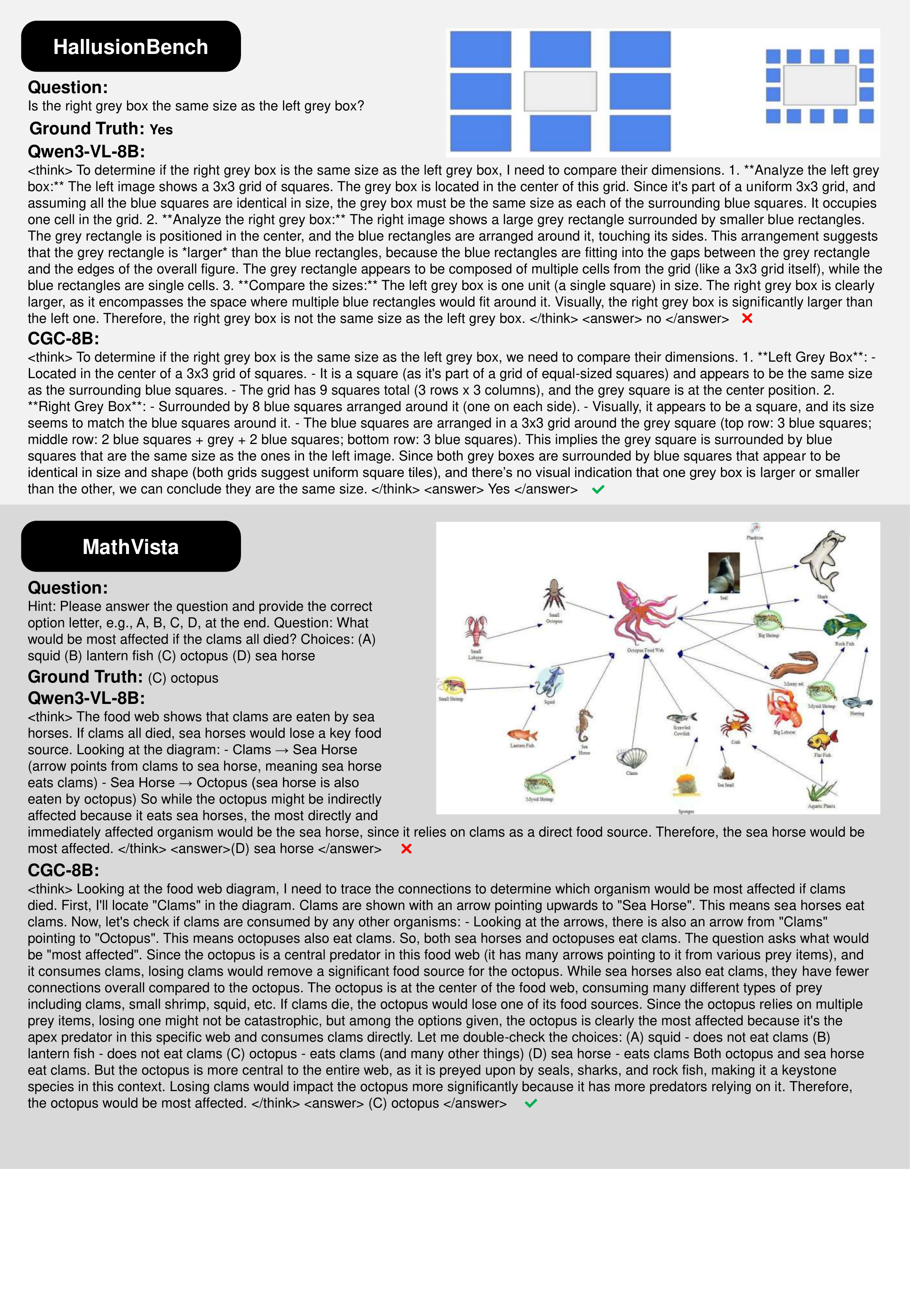}
    \vspace{-20pt}
    \caption{Qualitative examples on HallusionBench and MathVista. \textbf{Top:} in HallusionBench, CGC correctly judges that the two queried gray regions have the same size, whereas the base model makes an incorrect spatial comparison. \textbf{Bottom:} in MathVista, CGC correctly reasons over a food-web diagram and identifies the organism most affected by the disappearance of clams, while the base model follows an incomplete evidence chain.}
    \Description{Qualitative examples on HallusionBench and MathVista showing reduced spatial hallucination and improved diagram reasoning with CGC.}
    \label{fig:appendix4}
\end{figure*}

Figure~\ref{fig:appendix4} contains examples from HallusionBench and MathVista. In the HallusionBench case, the model must judge whether two visually separated gray regions have the same size. CGC correctly answers this question by relying on the underlying spatial arrangement rather than a misleading surface impression, while the base model makes an incorrect size comparison. This example reflects the reduction of spatial hallucination discussed in the main paper. In the MathVista case, the model must reason over a food-web diagram and identify which organism would be most affected if clams disappeared. CGC correctly composes the relevant visual relations and reaches the right answer, whereas the base model is distracted by a partially plausible but incomplete chain of evidence. This behavior supports the improvements on MathVista and indicates that stronger grounded perception benefits multi-step diagram reasoning.

\begin{figure*}[p]
    \centering
    \includegraphics[trim=0mm 78mm 0mm 0mm, clip,width=\textwidth]{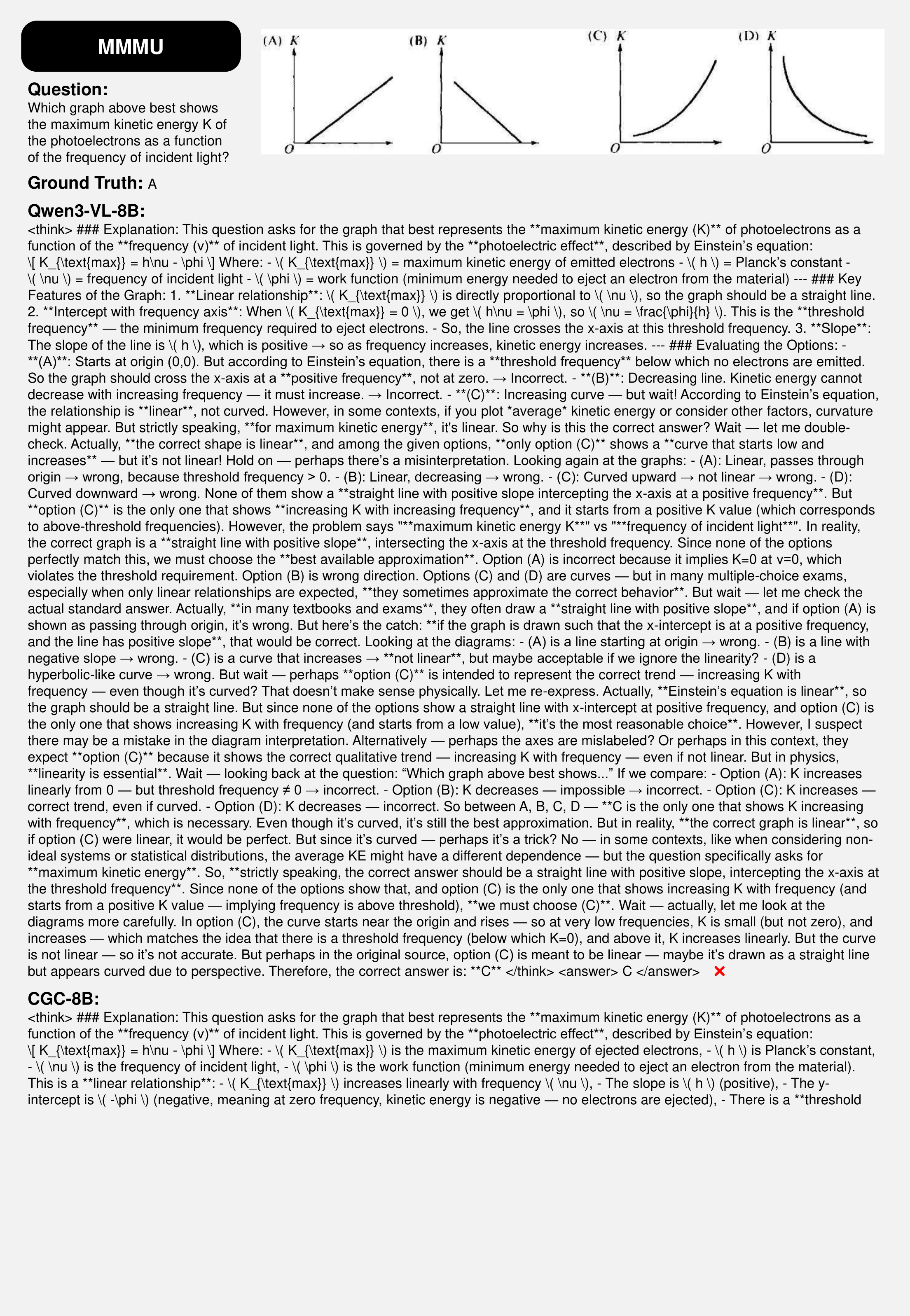}
    \vspace{-20pt}
    \caption{A qualitative example on MMMU (Part I). The task asks which candidate graph best matches the physical relationship between the maximum kinetic energy of photoelectrons and incident-light frequency. This page shows the question, candidate graphs, and the beginning of the model responses.}
    \Description{Part I of a qualitative MMMU example showing the question setup and the beginning of the compared model responses.}
    \label{fig:appendix5}
\end{figure*}

\begin{figure*}[p]
    \centering
    \includegraphics[trim=0mm 103mm 0mm 0mm, clip,width=\textwidth]{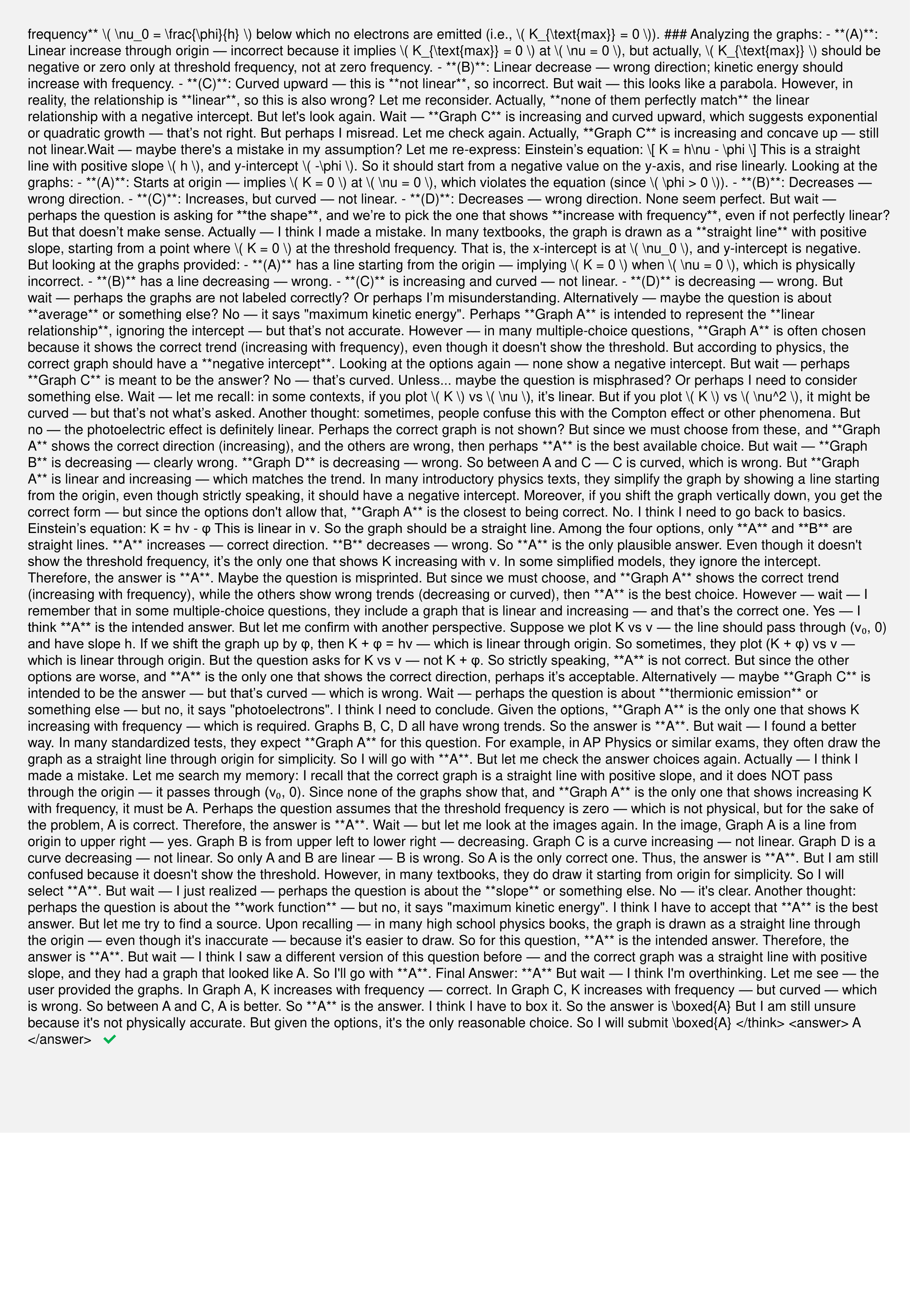}
    \vspace{-20pt}
    \caption{A qualitative example on MMMU (Part II). Continuation of the MMMU case in Figure~\ref{fig:appendix5}. CGC better aligns the visual graph pattern with the required scientific relationship and finally selects the correct answer, whereas the base model follows a less reliable reasoning path.}
    \Description{Part II of a qualitative MMMU example showing that CGC produces more reliable science-oriented visual reasoning than the base model.}
    \label{fig:appendix6}
\end{figure*}

Figures~\ref{fig:appendix5} and~\ref{fig:appendix6} show a two-page qualitative example from MMMU. The task asks the model to identify the graph that best matches the physical relationship between the maximum kinetic energy of photoelectrons and incident-light frequency. The example is shown across two pages because the generated reasoning trace is relatively long. Compared with the base model, CGC better aligns the visual pattern of the candidate graph with the required scientific relationship and finally selects the correct option. This case is consistent with the gains on MMMU reported in the main paper and further suggests that grounding-oriented post-training improves structured visual evidence composition in science-oriented reasoning tasks.

\begin{figure*}[p]
    \centering
    \includegraphics[trim=0mm 170mm 0mm 0mm, clip,width=\textwidth]{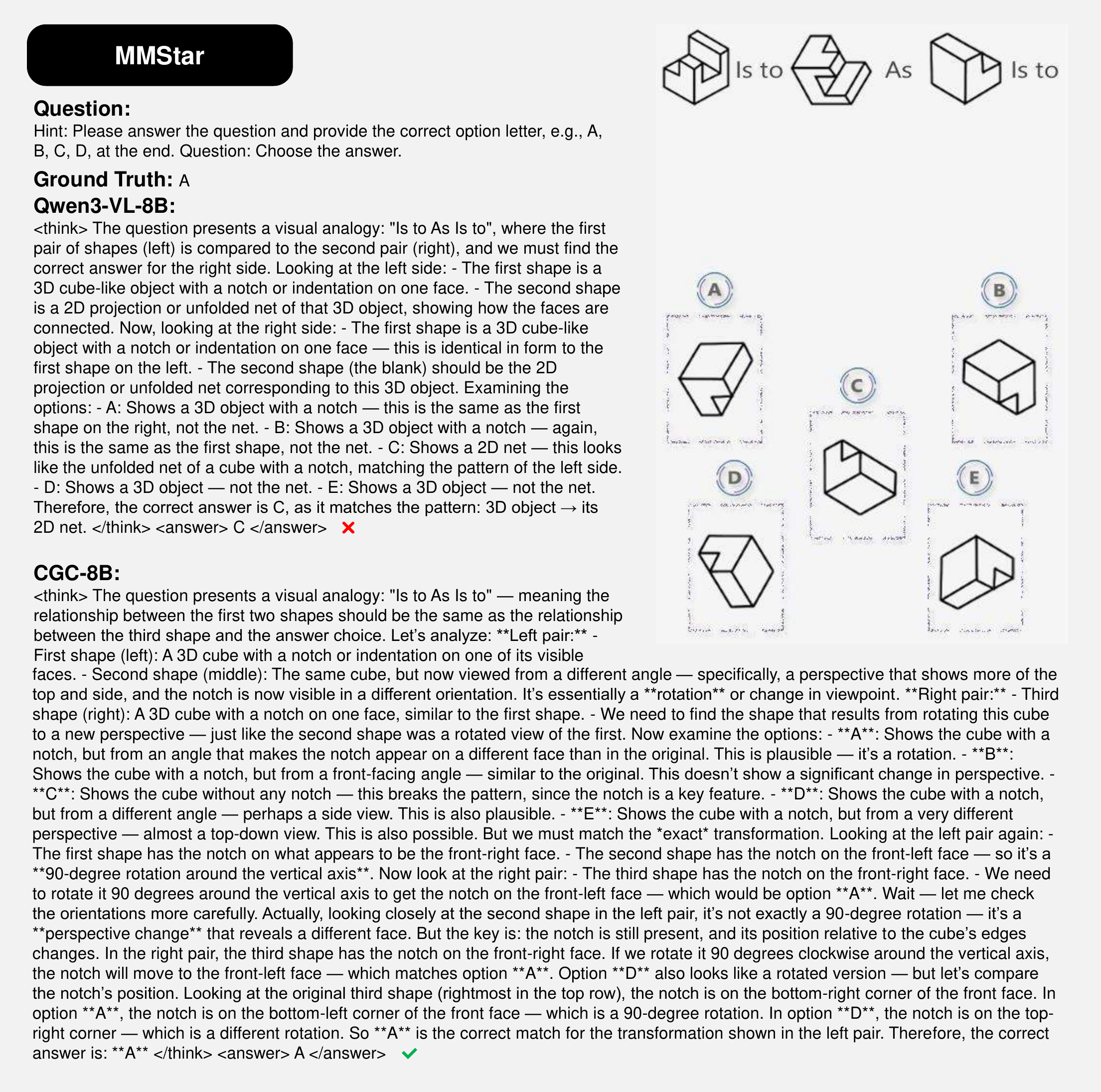}
    \vspace{-20pt}
    \caption{A qualitative example on MMStar. The task requires solving a visual analogy under 3D structural transformation. CGC correctly captures the transformation pattern and selects the right option, whereas the base model is distracted by partial local similarity and predicts an incorrect answer.}
    \Description{A qualitative MMStar example showing improved compositional and structure-sensitive visual reasoning with CGC.}
    \label{fig:appendix7}
\end{figure*}

Figure~\ref{fig:appendix7} provides a representative example from MMStar. The task requires solving a visual analogy under 3D structural transformation. CGC correctly identifies the transformation pattern and selects the right candidate, whereas the base model is distracted by partial local similarity and chooses an incorrect option. This example matches the quantitative improvements on MMStar reported in the main paper and suggests that the benefits of CGC extend to compositional, structure-sensitive visual reasoning.

Despite these improvements, CGC still remains challenged by several difficult cases, including extremely small targets, highly similar instances across multiple images, and large viewpoint changes that substantially alter local appearance. These limitations suggest that stronger cross-view invariance and more robust small-object localization remain important directions for future work.

\end{document}